\pdfoutput=1  %

\documentclass[10pt,twocolumn,letterpaper]{article}

\usepackage[pagenumbers]{cvpr}

\usepackage{cuted}      %
\usepackage{currfile}   %
\usepackage{caption}
\usepackage{booktabs}
\usepackage{array}
\usepackage{multirow}
\usepackage{amsmath}
\usepackage{amssymb}
\usepackage{xspace}
\usepackage{graphicx}
\usepackage{bm}

\newcommand{\paper}{EgoRayFormer\xspace}
\xspaceaddexceptions{\,\%]}
\DeclareMathOperator*{\argmin}{arg\,min}

\newcommand{\genfig}[2]{\includegraphics[width=#1]{#2}}

\definecolor{cvprblue}{rgb}{0.21,0.49,0.74}
\usepackage[pagebackref,breaklinks,colorlinks,allcolors=cvprblue]{hyperref}

\title{EgoRayFormer: Ray-Space Stereo Refinement for\\Egocentric 3D Human Pose Estimation}

\author{Haozhe Wang \qquad Songyuan Wu \qquad Jihai Xu \qquad Jie Zhao\\
Boden AI\\
{\tt\small \{haozhe.wang, songyuan.wu, jihai.xu, jie.zhao\}@bodenai.com}
}

\newcommand{\numAblEncN}{1\xspace}
\newcommand{\numAblEncPA}{$-$0.47\xspace}
\newcommand{\numAblEncPANeg}{3\xspace}
\newcommand{\numAblEncS}{+0.05\xspace}
\newcommand{\numAblEncVE}{$-$0.24\xspace}
\newcommand{\numAblEncVENeg}{3\xspace}
\newcommand{\numAblEncsd}{1.84\xspace}
\newcommand{\numAblGeo}{0.92\xspace}
\newcommand{\numAblGeoLow}{2.34\xspace}

\newcommand{\numAblGeoLowsd}{0.91\xspace}

\newcommand{\numAblGeosd}{0.51\xspace}

\newcommand{\numAblPriorN}{2\xspace}

\newcommand{\numAblPriorS}{$-$0.54\xspace}

\newcommand{\numAblRes}{2.83\xspace}

\newcommand{\numAblRessd}{0.78\xspace}

\newcommand{\numAblTrajN}{2\xspace}

\newcommand{\numAblTrajS}{$-$0.04\xspace}

\newcommand{\numEffEncShare}{91\xspace}

\newcommand{\numEffGflops}{106.9\xspace}
\newcommand{\numEffMsHi}{43.7\xspace}
\newcommand{\numEffMsLo}{42.8\xspace}
\newcommand{\numEffParams}{35.1\xspace}
\newcommand{\numEffResNetGflops}{17.9\xspace}
\newcommand{\numEffResNetMs}{35.4\xspace}
\newcommand{\numEffTsrGflops}{17.4\xspace}
\newcommand{\numEffTsrMs}{11.2\xspace}

\newcommand{\numFidHi}{74.10\xspace}
\newcommand{\numFidLo}{62.91\xspace}
\newcommand{\numFidMean}{68.01\xspace}
\newcommand{\numFidPA}{59.46\xspace}
\newcommand{\numFidPAref}{56.03\xspace}
\newcommand{\numFidRef}{65.58\xspace}

\newcommand{\numGapOwn}{20.62\xspace}

\newcommand{\numGapOwnsd}{3.12\xspace}
\newcommand{\numGapPA}{12.88\xspace}

\newcommand{\numGapPAsd}{3.71\xspace}
\newcommand{\numGapRA}{22.04\xspace}

\newcommand{\numGapRAsd}{4.23\xspace}
\newcommand{\numGapTestOwn}{30.91\xspace}
\newcommand{\numGapTestOwnN}{3\xspace}
\newcommand{\numGapTestOwnsd}{6.26\xspace}
\newcommand{\numGapTestPA}{17.87\xspace}
\newcommand{\numGapTestPAN}{3\xspace}
\newcommand{\numGapTestPAsd}{4.27\xspace}
\newcommand{\numGapTestRA}{34.02\xspace}
\newcommand{\numGapTestRAN}{3\xspace}
\newcommand{\numGapTestRAsd}{5.51\xspace}

\newcommand{\numLitTsrFourRW}{62.8\xspace}
\newcommand{\numLitTsrFourRWPA}{55.3\xspace}

\newcommand{\numRArw}{55.56\xspace}

\newcommand{\numRWTestFrames}{24,012\xspace}
\newcommand{\numRWTrainFrames}{79,705\xspace}
\newcommand{\numRWTrainSeq}{402\xspace}
\newcommand{\numRWValFrames}{30,083\xspace}
\newcommand{\numRWValSeq}{96\xspace}
\newcommand{\numRWonlyBest}{104.14\xspace}

\newcommand{\numRWonlyCIhi}{51.92\xspace}
\newcommand{\numRWonlyCIlo}{47.25\xspace}
\newcommand{\numRWonlyMPJPE}{49.55\xspace}

\newcommand{\numRWonlyPA}{40.59\xspace}
\newcommand{\numRWonlyRed}{52\xspace}
\newcommand{\numRWueBest}{72.89\xspace}

\newcommand{\numRWueCIhi}{50.11\xspace}
\newcommand{\numRWueCIlo}{45.04\xspace}
\newcommand{\numRWueMPJPE}{47.57\xspace}

\newcommand{\numRWueRed}{35\xspace}

\newcommand{\numSBdiff}{9.09\xspace}
\newcommand{\numSBdiffHi}{13.46\xspace}
\newcommand{\numSBdiffLo}{4.82\xspace}

\newcommand{\numSBours}{68.76\xspace}

\newcommand{\numSBpaDiff}{9.23\xspace}
\newcommand{\numSBtsr}{98.92\xspace}
\newcommand{\numSBtsrDiff}{30.16\xspace}
\newcommand{\numSBtsrDiffHi}{35.38\xspace}
\newcommand{\numSBtsrDiffLo}{25.45\xspace}

\newcommand{\numSBueDiff}{4.76\xspace}
\newcommand{\numSBueDiffHi}{5.09\xspace}
\newcommand{\numSBueDiffLo}{4.44\xspace}

\newcommand{\numSBueOurs}{25.57\xspace}
\newcommand{\numSBueOursPA}{23.85\xspace}

\newcommand{\numSBueTsr}{30.33\xspace}
\newcommand{\numSBueTsrPA}{27.13\xspace}

\newcommand{\numSessSeenHi}{55.97\xspace}
\newcommand{\numSessSeenLo}{44.72\xspace}
\newcommand{\numSessUnseen}{47.99\xspace}

\newcommand{\numTSRvalDiff}{23.24\xspace}
\newcommand{\numTSRvalDiffsd}{2.62\xspace}
\newcommand{\numTSRvalWorse}{3\xspace}
\newcommand{\numUEtwoBest}{30.53\xspace}

\newcommand{\numUEtwoCIhi}{18.55\xspace}
\newcommand{\numUEtwoCIlo}{16.80\xspace}
\newcommand{\numUEtwoMPJPE}{17.64\xspace}

\newcommand{\numUEtwoRed}{42\xspace}
\newcommand{\numUEtwoTestFrames}{123,480\xspace}
\newcommand{\numUEtwoTrainFrames}{1,002,671\xspace}
\newcommand{\numUEtwoTrainSeq}{12,139\xspace}
\newcommand{\numUEtwoValFrames}{127,937\xspace}
\newcommand{\numUEtwoValSeq}{1,545\xspace}

\newcommand{\numAnaAlongFarHi}{48.9\xspace}
\newcommand{\numAnaAlongFarLo}{45.7\xspace}
\newcommand{\numAnaFarAlong}{46.1\xspace}
\newcommand{\numAnaFarLateral}{85.4\xspace}
\newcommand{\numAnaFarMPJPE}{96.4\xspace}
\newcommand{\numAnaGateDropA}{0.19\xspace}
\newcommand{\numAnaGateDropB}{0.01\xspace}
\newcommand{\numAnaGateDropC}{0.24\xspace}
\newcommand{\numAnaGateMax}{0.56\xspace}
\newcommand{\numAnaGateMeanA}{0.42\xspace}
\newcommand{\numAnaGateMeanB}{0.30\xspace}
\newcommand{\numAnaGateMeanC}{0.34\xspace}
\newcommand{\numAnaGateRiseA}{0.24\xspace}
\newcommand{\numAnaGateRiseB}{0.03\xspace}
\newcommand{\numAnaGateRiseC}{0.27\xspace}
\newcommand{\numAnaNearMPJPE}{13.5\xspace}
\newcommand{\numAnaValidTwo}{99.8\xspace}
\newcommand{\numAnaXsGainMax}{2.6\xspace}

\begin{document}
\twocolumn[{%
  \renewcommand\twocolumn[1][]{#1}%
  \maketitle
  \begin{center}
    \captionsetup{type=figure}
    \genfig{\textwidth}{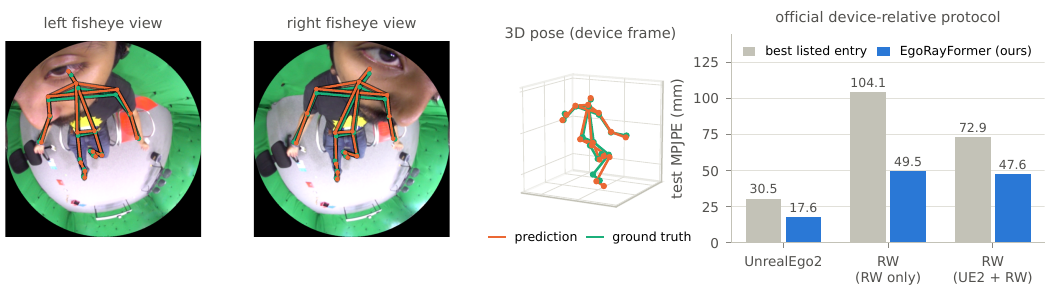}
    \captionof{figure}{\textbf{Device-relative body pose from stereo fisheye images.}
      Left: prediction and ground truth for an UnrealEgo-RW validation frame selected at the median error.
      Right: our local three-seed test MPJPE and published leaderboard values. These compare complete systems;
      correspondence between the RW leaderboard split and the current public split remains unverified
      (\cref{sec:exp_main}).}
    \label{fig:teaser}
  \end{center}
  \vspace{1mm}
}]

\begin{abstract}
Head-mounted stereo fisheye cameras provide a compact view of the wearer's body, but strong distortion, partial
visibility and a short baseline make 3D pose estimation difficult. Existing refinement decoders sample image features
independently in each view, without requiring the samples to correspond to a common 3D hypothesis. We introduce
\paper, a causal pose estimator that couples stereo feature sampling and geometric refinement in ray space.
For each joint, the decoder projects shared 3D hypotheses along and around its viewing ray into both images.
A geometric branch then corrects the projected joint locations and intersects the resulting rays; a learned gate
blends this geometric estimate with a learned pose update. This formulation introduces an explicit stereo constraint
while retaining flexibility when the rays provide weak depth constraints.
In local evaluations on the official device-relative test splits, \paper achieves three-seed mean MPJPEs of
\numUEtwoMPJPE\,mm on UnrealEgo2 and \numRWonlyMPJPE\,mm on UnrealEgo-RW, improving to \numRWueMPJPE\,mm with
UnrealEgo2 pre-training. In paired validation ablations, removing the geometric branch increases MPJPE in all three
seeds at both input resolutions. With a common ResNet-18 encoder and input resolution, our system reduces RW test
MPJPE by \numSBdiff\,mm relative to the device-relative EgoPoseFormer baseline. A separate protocol analysis
quantifies the difference between device-relative and oracle-root evaluation.
\end{abstract}

\section{Introduction}
\label{sec:intro}

Estimating the wearer's body pose from headset-mounted cameras enables avatar animation and interaction without
external cameras~\cite{rhodin2016egocap,xu2019mo2cap2,tome2019xregopose,jiang2022avatarposer}. A pair of downward-looking
fisheye cameras offers wide coverage in a compact rig~\cite{zhao2021egoglass,akada2022unrealego,akada2024stereo}, but
the resulting images are difficult to interpret. Limbs appear foreshortened, distal joints occupy few pixels, and body
parts are frequently occluded or outside the field of view. The short stereo baseline adds a geometric difficulty:
for cameras about 12\,cm apart, the rays to a joint 1.5\,m away meet at an angle of only about five degrees.
Accurate reconstruction therefore requires both useful image evidence and a way to handle weak geometric constraints.

Recent pose decoders refine a coarse 3D estimate by sampling features near its image
projections~\cite{yang2024egoposeformer,li2026egoposeformerv2,azam2026tsrego}. This design concentrates computation around
the body, but the sampling offsets are predicted independently for each view. The sampled image locations need not
be projections of a common 3D hypothesis, leaving the relation between the two views implicit in feature fusion.
Direct triangulation provides an explicit relation, yet small localisation errors can produce large depth errors
when the viewing rays are nearly parallel. These observations motivate a refinement step that connects stereo
sampling to 3D geometry while retaining a learned alternative to the geometric estimate.

We propose \paper, a causal decoder for stereo fisheye pose estimation. Each refinement layer places shared 3D
hypotheses in a local coordinate frame aligned with the ray from the device to a joint. Projecting each hypothesis
into both cameras couples the locations at which the two feature maps are sampled. From these features, a geometric
branch predicts per-view corrections, computes a differentiable intersection of the corrected rays, and blends the
result with a learned pose update. The complete model combines this refinement with a stereo image encoder and causal
temporal attention, and predicts joint coordinates relative to the device without ground-truth inputs at inference.

Our evaluation distinguishes system performance from evidence about individual components. On the official test
splits, local three-seed evaluations give MPJPEs of \numUEtwoMPJPE\,mm on UnrealEgo2, \numRWonlyMPJPE\,mm on
UnrealEgo-RW with RW-only training, and \numRWueMPJPE\,mm with UnrealEgo2 pre-training. With ResNet-18 and matched
input resolution, \paper has a \numSBdiff\,mm lower RW test MPJPE than our device-relative adaptation of the official
EgoPoseFormer code. This compares complete systems, including their temporal context and training recipes.
More specifically, removing the geometric branch increases validation MPJPE by \numAblGeo\,mm at $384^2$ and
\numAblGeoLow\,mm at $256^2$, with the same direction in every paired seed. These ablations support the branch as a
whole.

Evaluation coordinates also matter. Device-relative estimation requires the model to recover body position relative
to the headset, whereas root-relative scoring removes global body translation. Some pipelines additionally use the
root to project decoder anchors into the images. We therefore state the inference inputs and scoring coordinates
for each comparison, and measure the complete device/oracle-root protocol gap using one architecture.
Published leaderboard values provide external context, with the unresolved RW split correspondence stated alongside
the comparison (\cref{sec:exp_main}). Earlier development use of the public test sets is disclosed in
\cref{sec:exp_setup,app:protocol}.

Our contributions are:
\begin{itemize}
  \item A ray-space refinement decoder that couples stereo sampling through shared 3D hypotheses and combines
    corrected-ray intersection with a learned pose update.
  \item A multi-seed evaluation on UnrealEgo2 and UnrealEgo-RW, including comparisons at a common backbone and
    resolution, paired ablations of the geometric branch, and analysis of residual errors and gate behaviour.
  \item A paired analysis of device-relative and oracle-root protocols that quantifies their combined effect for
    a fixed architecture and clarifies the interpretation of cross-protocol results.
\end{itemize}

\section{Related Work}
\label{sec:related}

\paragraph{Egocentric body pose estimation.}
Early systems estimated body pose from helmet- or cap-mounted fisheye
cameras~\cite{rhodin2016egocap,xu2019mo2cap2}, followed by headset-mounted monocular
methods~\cite{tome2019xregopose,tome2020selfpose}, scene-aware models~\cite{wang2023sceneego}, and fisheye-specific
representations~\cite{wang2024fisheyevit}. Stereo approaches such as EgoGlass~\cite{zhao2021egoglass},
UnrealEgo~\cite{akada2022unrealego}, Ego3DPose~\cite{kang2023ego3dpose} and EgoTAP~\cite{kang2024egotap} lift per-view
2D heatmaps to 3D. Akada \etal~\cite{akada2024stereo} introduced UnrealEgo2 and UnrealEgo-RW and used scene depth
from structure-from-motion to support video-based estimation. Rear cameras~\cite{akada2025egorear} and headset
pose~\cite{camiletto2025frame} provide additional observations in other capture settings.

Our decoder builds on the proposal-and-refinement design of EgoPoseFormer~\cite{yang2024egoposeformer}, whose
deformable attention samples features around projected joints. EgoPoseFormer v2~\cite{li2026egoposeformerv2} adds
causal temporal attention, per-joint uncertainty and auto-labelling for a four-camera headset.
TSR-Ego~\cite{azam2026tsrego} combines causal mixing of stereo feature maps with temporal, joint and per-view
deformable attention. We instead tie the sampling locations across views to shared 3D hypotheses and use corrected-ray
intersection within the refinement step. This distinction concerns the spatial update; causal attention and the
coarse-to-fine architecture have precedents in these methods.

\paragraph{Geometry-guided multi-view refinement.}
Learnable triangulation~\cite{iskakov2019learnable} combines calibrated views through algebraic or volumetric
triangulation. MvP~\cite{wang2021mvp} uses projective attention, while MVGFormer~\cite{liao2024mvgformer} alternates
learned 2D corrections and confidences with weighted triangulation. In multi-camera detection,
PETR~\cite{liu2022petr} encodes 3D position and BEVFormer~\cite{li2022bevformer} projects reference points into camera
views. Sparse4D~\cite{lin2022sparse4d} projects shared 3D keypoints into multiple feature maps, providing a direct
precedent for our shared-hypothesis sampling; RayDN~\cite{liu2024raydn} uses ray-based samples for denoising.
\paper adapts shared spatial hypotheses and geometric refinement to a two-camera fisheye rig. The hypotheses are
organised in a ray-local frame, and the geometric estimate is blended with a learned update to accommodate the
weak depth constraints of the short baseline.

\paragraph{Fisheye and camera-aware image features.}
FisheyeViT~\cite{wang2024fisheyevit} represents fisheye images using undistorted tangent-plane patches. We adopt
this tokenisation with ray-direction inputs and stereo cross-view blocks inspired by CroCo and
DUSt3R~\cite{weinzaepfel2022croco,wang2024dust3r}, using a DINOv3 backbone~\cite{simeoni2025dinov3,oquab2024dinov2}.
Other camera-aware transformers incorporate geometry through relative positional
encoding~\cite{miyato2024gta,li2025prope}. We use absolute viewing-ray channels and evaluate the additional encoder
components together against a plain DINOv3 encoder (\cref{sec:ablation}).

\paragraph{Evaluation coordinates and inference inputs.}
The UnrealEgo2 and UnrealEgo-RW benchmarks~\cite{akada2024stereo} evaluate device-relative joint coordinates.
Root-relative scoring, also common in third-person pose estimation~\cite{ionescu2014human36m}, removes global body
translation. This scoring choice is distinct from supplying a root position to the model. The public EgoPoseFormer
implementation for UnrealEgo uses the ground-truth pelvis as the projection origin; TSR-Ego reports root-relative
results without specifying whether its inference-time root is predicted or ground truth. We keep these distinctions
explicit and use a paired protocol experiment to assess their combined effect (\cref{sec:protocol}).

\section{Method}
\label{sec:method}

\begin{figure*}[t]
  \centering
  \genfig{\linewidth}{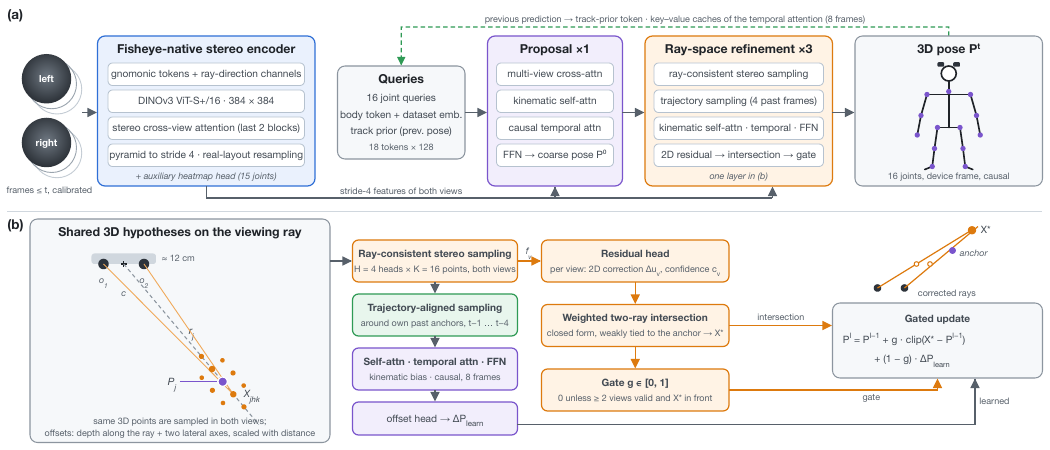}
  \caption{\textbf{Overview of \paper.} A stereo encoder provides image features for a coarse pose proposal and three
    refinement layers. Each refinement layer projects shared 3D hypotheses into both views, aggregates their sampled
    features, and combines a learned pose update with a regularised ray-intersection update. Causal attention and
    cached image features provide temporal context for frame-by-frame inference.}
  \label{fig:arch}
\end{figure*}

\paper estimates body pose from a causal stream of stereo fisheye images (\cref{fig:arch}). Its decoder uses a shared
3D hypothesis as the unit of stereo feature aggregation: each hypothesis is projected into both calibrated views, and
the paired features support pose refinement. A second path predicts image-plane corrections, intersects the resulting
viewing rays, and blends this geometric update with a learned update. We first describe this decoder, then the encoder,
temporal context and training objective.

\subsection{Problem setup}
\label{sec:setup}

A head-mounted rig carries two downward-looking fisheye cameras $v \in \{1, 2\}$ about 12\,cm apart. Each camera uses
the polynomial omnidirectional calibration of Scaramuzza \etal~\cite{scaramuzza2006toolbox}, with known pose relative to
the device. We denote by $\pi_v(X)$ the normalised image projection of device-frame point $X$, and by $(o_v,d_v(u))$
the camera centre and unit viewing ray through image point $u$, expressed in the same frame. The device centre is
$c=(o_1+o_2)/2$.

At frame $t$, the model receives images $I^t_1,I^t_2$ and predicts $J{=}16$ joint positions in the device frame, following
the official UnrealEgo2 and UnrealEgo-RW benchmarks~\cite{akada2024stereo}. Inference uses current and past images and
camera calibration, without ground-truth root position, scene depth or headset pose. We write $P^{t,l}\in\mathbb{R}^{J\times3}$ for the pose
after decoder stage $l$, where $l=0$ is the proposal and $l=1,\ldots,L$ are the $L{=}3$ refinement stages. Below we omit
$t$ and use $P_j=P^{l-1}_j$ for the anchor of joint $j$ at a refinement stage.

\subsection{Queries and coarse proposal}
\label{sec:proposal}

The encoder (\cref{sec:encoder}) produces a stride-4 feature map with $C{=}128$ channels for each view. The decoder
contains $J$ learned joint queries, a body token with a learned dataset embedding, and a track-prior token that embeds
the previous predicted pose. A learned null token is used at the start of a sequence. During training, the previous
ground-truth pose is perturbed by Gaussian noise ($\sigma=3$\,cm) and replaced by the null token with probability $0.3$.

The proposal layer applies camera-conditioned cross-attention to both feature maps, followed by token self-attention,
causal temporal attention and a feed-forward layer. Cross-attention adds a learned camera embedding to each query and
linearly fuses the two view outputs~\cite{li2026egoposeformerv2}. Self-attention uses a learned per-head bias between joints indexed
by skeleton hop distance. A regression head maps the joint tokens to the coarse pose $P^0$.

\subsection{Ray-consistent stereo sampling}
\label{sec:sampling}

For joint query $q_j$ and anchor $P_j$, define the ray direction $r_j=(P_j-c)/\lVert P_j-c\rVert$. Together with two
orthogonal lateral axes $e_{j,1}$ and $e_{j,2}=r_j\times e_{j,1}$, it forms a local coordinate frame. The first lateral
axis is perpendicular to the device vertical axis and $r_j$; a fixed alternative axis handles their near-parallel case.
Each of $H{=}4$ heads predicts $K{=}16$ offsets $\delta_{jhk}\in\mathbb{R}^3$ from $q_j$, in centimetres at a reference
distance of 50\,cm. The resulting hypotheses are
\begin{equation}
  X_{jhk}=P_j+s_j\bigl(\delta^r_{jhk}r_j+\delta^1_{jhk}e_{j,1}+\delta^2_{jhk}e_{j,2}\bigr),
  \label{eq:hypotheses}
\end{equation}
where $s_j=\operatorname{clip}(\lVert P_j-c\rVert/50\,\mathrm{cm},0.5,4)$. The radial component varies depth and the
lateral components cover the surrounding region. Within the unclipped distance range, scaling approximately preserves
the lateral angular footprint. Offsets are initialised to $\pm3$\,cm radially and to lateral rings of 3 and 6\,cm,
rotated across heads.

Each hypothesis is projected into both views, $u_{v,jhk}=\pi_v(X_{jhk})$. Let $m_{v,jhk}$ indicate whether it projects
inside the valid image domain. Bilinear sampling of the value-projected feature maps gives
$f_{v,jhk}=V^h_v(u_{v,jhk})$; invalid samples are replaced by a learned missing-view vector. We fuse the paired features
and their validity flags at each hypothesis:
\begin{equation}
  z_{jhk}=\operatorname{GELU}\bigl(W_f[\,f_{1,jhk};m_{1,jhk};f_{2,jhk};m_{2,jhk}\,]\bigr).
  \label{eq:fusion}
\end{equation}
The projections share a 3D hypothesis; this construction does not ensure that the hypothesised point is occupied or
visible in both images. Query-predicted weights $a_{jhk}=\operatorname{softmax}_k(W_aq_j)_{hk}$ aggregate the fused
features into $\Delta q_j=W_o\bigoplus_h\sum_k a_{jhk}z_{jhk}$, which is added to the query. The same weights produce
per-view features $\bar f_{v,j}=\bigoplus_h\sum_k a_{jhk}f_{v,jhk}$ for geometric refinement. Here $\bigoplus$ denotes
concatenation across heads.

In the deformable stereo attention of EgoPoseFormer and TSR-Ego~\cite{yang2024egoposeformer,azam2026tsrego}, each view
predicts 2D offsets around the projected anchor. Samples with the same offset index need not be projections of a common
3D hypothesis. Our construction follows the shared 3D sampling principle of Sparse4D~\cite{lin2022sparse4d}, with offsets
expressed in a ray-local frame for a short-baseline fisheye rig. Gradients are stopped through the anchor coordinates
used for projection and sampling; the residual pose updates retain their gradient path between refinement layers.

\subsection{Geometric refinement}
\label{sec:geometry}

After sampling, token self-attention, causal temporal attention and a feed-forward layer update the tokens. An offset
head predicts $\Delta P^{\mathrm{learn}}_j$. In parallel, a residual head uses the per-view features and updated query
$q'_j$ to predict image-plane corrections and confidence logits, following the appearance-based refinement of
MVGFormer~\cite{liao2024mvgformer}:
\begin{equation}
\begin{aligned}
  (\Delta u^{\mathrm{px}}_{v,j},\tilde c_{v,j})&=\operatorname{MLP}([\,\bar f_{v,j};q'_j\,]),\\
  u'_{v,j}&=\operatorname{clip}_{[0,1]^2}\bigl(\bar u_{v,j}+D\Delta u^{\mathrm{px}}_{v,j}\bigr).
\end{aligned}
  \label{eq:residual}
\end{equation}
Here $\bar u_{v,j}=\operatorname{clip}_{[0,1]^2}(\pi_v(P_j))$ is the anchor sampling coordinate, and
$D=\operatorname{diag}(1/W_{\mathrm{feat}},1/H_{\mathrm{feat}})$ converts corrections from feature pixels to normalised image coordinates for a
feature map of width $W_{\mathrm{feat}}$ and height $H_{\mathrm{feat}}$. The corrected points define rays $(o_v,d_{v,j})=(o_v,d_v(u'_{v,j}))$.
Let $m_{v,j}$ indicate the validity of
the anchor projection, and let $w_{v,j}=\sigma(\tilde c_{v,j})m_{v,j}$, where $\sigma$ is the sigmoid function. The
geometric candidate minimises a confidence-weighted ray distance with an anchor regulariser:
\begin{equation}
\begin{aligned}
  X^*_j=\argmin_X\;&\sum_v w_{v,j}\bigl\lVert(I-d_{v,j}d_{v,j}^{\top})(X-o_v)\bigr\rVert^2\\
  &+\epsilon\lVert X-P_j\rVert^2.
\end{aligned}
\label{eq:intersection}
\end{equation}
This differentiable operation solves a $3\times3$ linear system. We use $\epsilon=2\cdot10^{-5}$ to make the system
invertible when the rays provide insufficient depth information. The regulariser draws weakly constrained directions
towards the anchor; its effect depends on both the ray angle and the predicted weights.

The geometric candidate is valid when the anchor projects inside both views and $X^*_j$ has positive depth along
both rays. A learned scalar gate $g_j\in[0,1]$, set to zero for invalid candidates, combines the two updates:
\begin{equation}
  P^l_j=P^{l-1}_j+g_j\operatorname{clip}\bigl(X^*_j-P^{l-1}_j\bigr)
      +(1-g_j)\Delta P^{\mathrm{learn}}_j,
  \label{eq:gate}
\end{equation}
where clipping limits each coordinate of the geometric update to $\pm60$\,cm. The gate receives view-confidence and
heatmap-confidence summaries, projection and intersection validity, ray-angle information, and a geometric covariance
summary computed under an assumed angular noise corresponding to half a feature pixel. These inputs describe the
observation geometry; the gate is trained as a blending weight. Its output bias is initialised to $-2$
($g\approx0.12$), and the residual head starts at zero correction and confidence $0.5$. Auxiliary 2D and 3D losses
supervise the geometric branch (\cref{sec:training}); \cref{sec:analysis} examines the learned gate behaviour.

\subsection{Fisheye-native stereo encoder}
\label{sec:encoder}

The encoder adapts DINOv3 ViT-S+/16~\cite{simeoni2025dinov3,dosovitskiy2021vit} to $384\times384$ stereo fisheye images.
Following FisheyeViT~\cite{wang2024fisheyevit}, a precomputed calibration grid resamples each image into virtual
$16\times16$ tangent-plane patches. Three additional channels encode the device-frame viewing ray at each virtual
pixel; their patch-embedding weights start at zero. Two cross-view attention blocks, placed after the last two ViT
blocks, exchange features between cameras with zero-initialised layer scales~\cite{weinzaepfel2022croco,wang2024dust3r}.
A simple feature pyramid~\cite{li2022vitdet} produces stride-4 features with 128 channels. An inverse layout map resamples
them into the original fisheye image coordinates for calibrated decoder sampling. An auxiliary head predicts
$64\times64$ heatmaps for the 15 joints other than the head, providing encoder supervision and heatmap-confidence inputs
to the gate. \Cref{sec:ablation} evaluates these encoder adaptations as a group.

\subsection{Causal temporal modelling}
\label{sec:temporal}

\paragraph{Token history.} Every decoder layer applies temporal attention separately to each token, using the current
frame and up to seven preceding frames, with rotary frame embeddings~\cite{su2024roformer,li2026egoposeformerv2}.
Training processes clips of eight frames with a causal band mask. At inference, per-layer key--value caches support
frame-by-frame execution. With identical images, track priors and history, the streaming and parallel causal passes
agree within $10^{-3}$\,cm in our unit tests. The training prior uses perturbed ground truth, whereas inference feeds
back the previous prediction (\cref{sec:proposal}).

\paragraph{Trajectory-aligned feature sampling.} A query also samples the four preceding frames. For each past frame,
the offsets of \cref{eq:hypotheses} are placed around that joint's anchor from the preceding decoder stage and projected
using that frame's calibration. Query-predicted weights with a learned age bias are normalised jointly over past ages
and sampling points. The aggregated features are added to the current-frame sampling output; past value maps are
cached during streaming. \Cref{sec:ablation} evaluates this component separately.

\subsection{Training}
\label{sec:training}

We first train the encoder on 2D heatmaps at $256\times256$ using the target dataset's training split, then train the
full model end to end at $384\times384$ on eight-frame clips. The objective is
\begin{equation}
\begin{aligned}
  \mathcal{L}={}&\lambda_{\mathrm p}\mathcal{L}_{3\mathrm D}(P^0)
    +\sum_{l=1}^{L}\lambda_l\mathcal{L}_{3\mathrm D}(P^l)+\lambda_{\mathrm{hm}}\mathcal{L}_{\mathrm{hm}}\\
  &+\lambda_{\mathrm{rep}}\mathcal{L}_{\mathrm{rep}}+\lambda_{2\mathrm D}\mathcal{L}_{2\mathrm D}
    +\lambda_{\mathrm{geo}}\mathcal{L}_{\mathrm{geo}}+\lambda_{\mathrm{bone}}\mathcal{L}_{\mathrm{bone}}\\
  &+\lambda_{\mathrm{vel}}\mathcal{L}_{\mathrm{vel}}+\lambda_{\mathrm{jerk}}\mathcal{L}_{\mathrm{jerk}}
    +\lambda_{\mathrm{dn}}\mathcal{L}_{\mathrm{dn}}.
\end{aligned}
\label{eq:loss}
\end{equation}
The pose terms are mean Euclidean joint errors, and $\mathcal{L}_{\mathrm{hm}}$ supervises the heatmaps.
$\mathcal{L}_{\mathrm{rep}}$ measures final-pose reprojection error. The geometric branch losses supervise corrected 2D
points ($\mathcal{L}_{2\mathrm D}$) and ray-intersection candidates for joints annotated in both views
($\mathcal{L}_{\mathrm{geo}}$), averaging each loss over refinement layers. The remaining terms supervise bone lengths
and left--right symmetry, match temporal first differences to ground truth, and penalise the predicted third temporal
difference. One group of denoising queries~\cite{li2022dndetr} uses ground-truth poses perturbed by 5\,cm Gaussian noise
as additional anchors, with $\mathcal{L}_{\mathrm{dn}}$ supervising their reconstruction through the refinement layers.
The weights are $\lambda_{\mathrm p}=\lambda_l=1$, $\lambda_{\mathrm{hm}}=10$,
$\lambda_{\mathrm{rep}}=\lambda_{2\mathrm D}=0.05$, $\lambda_{\mathrm{geo}}=0.5$, $\lambda_{\mathrm{bone}}=0.1$,
$\lambda_{\mathrm{vel}}=0.5$, $\lambda_{\mathrm{jerk}}=0.2$ and $\lambda_{\mathrm{dn}}=1$ (3D terms in cm).

We use AdamW~\cite{loshchilov2019adamw} with learning rate $10^{-3}$, weight decay $5\cdot10^{-4}$, 500 warm-up steps,
and a ViT learning-rate scale of 0.15 with layer-wise decay 0.8. Training runs for 12 epochs on UnrealEgo-RW, with
tenfold learning-rate reductions after epochs 8 and 10; UnrealEgo2 uses 3 epochs with reductions after epochs 2 and 2.5.
We use batches of 16 clips, bf16 mixed precision, gradient clipping at norm 5, and photometric augmentation shared across
views and frames. Evaluation uses an exponential moving average of the weights with decay 0.999. In the sequential
UnrealEgo2 $\to$ UnrealEgo-RW setting, RW training starts from the UnrealEgo2 model and follows the RW schedule.
Additional implementation details are given in \cref{app:implementation}.

\section{Experiments}
\label{sec:exp}

\begin{table*}[t]
  \centering
  \footnotesize
  \setlength{\tabcolsep}{4.5pt}
  \begin{tabular}{@{}llll cc cc cc@{}}
\toprule
 & & & & \multicolumn{2}{c}{UnrealEgo2} & \multicolumn{2}{c}{RW, trained on RW} & \multicolumn{2}{c}{RW, trained on UE2 + RW} \\
\cmidrule(lr){5-6}\cmidrule(lr){7-8}\cmidrule(lr){9-10}
Method & Backbone & Frames & Extra input & MPJPE & PA & MPJPE & PA & MPJPE & PA \\
\midrule
EgoGlass$^{\dagger}$~\cite{zhao2021egoglass} & R18 & 1 & -- & 79.64 & 58.22 & -- & -- & 97.86 & 69.92 \\
UnrealEgo~\cite{akada2022unrealego} & R18 & 1 & -- & 72.82 & 52.90 & -- & -- & 92.48 & 67.15 \\
Akada \etal~\cite{akada2024stereo} & R18 & 5 & scene depth & 30.53 & 26.72 & 104.14 & 82.18 & 72.89 & 56.19 \\
\midrule
\paper (ours) & ViT-S+ & stream & calibration & \textbf{17.64}\,{\scriptsize$\pm$0.12} & \textbf{16.86}\,{\scriptsize$\pm$0.08} & \textbf{49.55}\,{\scriptsize$\pm$0.40} & \textbf{40.59}\,{\scriptsize$\pm$0.30} & \textbf{47.57}\,{\scriptsize$\pm$0.34} & \textbf{37.21}\,{\scriptsize$\pm$0.46} \\
\bottomrule
\end{tabular}

  \caption{\textbf{Comparison with published leaderboard values} (device-relative protocol, test sets, mm,
    lower is better). Ours: local evaluations on the current official splits, mean\,$\pm$\,s.d.\ over three seeds;
    results have not been submitted to the leaderboard. The split version underlying the published RW values is
    unverified (\cref{sec:exp_main}). Backbone: R18 is ResNet-18 at $256^2$, ViT-S+ is DINOv3 ViT-S+/16 at
    $384^2$. Frames: input frames per prediction; stream denotes causal streaming over the whole sequence. Extra input:
    inputs used at inference besides the stereo images; scene depth is rendered from a structure-from-motion
    reconstruction of the scene~\cite{akada2024stereo}, calibration denotes the fisheye camera model and the stereo rig
    (\cref{sec:exp_setup}). $^{\dagger}$Re-implemented by the benchmark authors. Full comparison conditions are given
    in \cref{app:conditions}.}
  \label{tab:main}
\end{table*}

\subsection{Setup}
\label{sec:exp_setup}

\paragraph{Datasets.}
UnrealEgo2~\cite{akada2024stereo} is rendered with a synthetic stereo fisheye rig. Its official split has
\numUEtwoTrainSeq training sequences (\numUEtwoTrainFrames frames), \numUEtwoValSeq validation sequences
(\numUEtwoValFrames frames) and \numUEtwoTestFrames test frames. UnrealEgo-RW~\cite{akada2024stereo} (RW) was captured
with a physical head-mounted rig and has \numRWTrainSeq training sequences (\numRWTrainFrames frames), \numRWValSeq
validation sequences (\numRWValFrames frames) and \numRWTestFrames test frames. As on the leaderboard, RW results are
reported for models trained on RW only and for models trained on UnrealEgo2 and RW; for the latter, the RW model is
initialised with the UnrealEgo2 model of the same seed (\cref{sec:training}).

\paragraph{Calibration.}
UnrealEgo2 has a single synthetic rig. For RW we keep the intrinsics of the dataset's calibration files and fit the
camera-from-device transform of every capture session to the device-frame and camera-frame joint annotations of the
training split; validation and test annotations are never used. One test session does not occur in the training data
and uses the rig of the chronologically nearest session.

\paragraph{Protocol and metrics.}
Our primary evaluation follows the official device-relative protocol: predictions are the 3D positions of 16 joints
relative to the device centre, with no ground-truth quantity used at inference. We report MPJPE and PA-MPJPE (after
rigid alignment with scale) in mm, using a common scorer based on the public EgoPoseFormer evaluation code.
Ablations additionally report the mean per-joint velocity error MPJVE, the error of frame-to-frame joint displacements
in mm per frame. Every sequence is
processed in one causal stream from its first frame, in FP32.

\paragraph{Model selection and statistics.}
We train each configuration with three seeds and select each checkpoint by validation MPJPE. Before the final
evaluations reported here, we froze the checkpoints and evaluation setup; each checkpoint was then evaluated on the
test set once. We report the mean and sample standard deviation across seeds. Confidence intervals resample test
sequences with the fitted models held fixed (10{,}000 bootstrap samples, frame-weighted). Ablations and error analyses
use validation data. The test splits have been public since December 2025, and preliminary test evaluations informed
some development decisions. No result from those evaluations is reported here, but the test sets were therefore not
unseen during development. \Cref{app:protocol} details this history and the final evaluation procedure.

\subsection{Comparison with published benchmark results}
\label{sec:exp_main}

\paper achieves a mean MPJPE of \numUEtwoMPJPE\,mm on UnrealEgo2, \numRWonlyMPJPE\,mm on RW with RW-only training,
and \numRWueMPJPE\,mm on RW after UnrealEgo2 pre-training (\cref{tab:main}). Relative to the lowest published values
listed in the table (\numUEtwoBest, \numRWonlyBest and \numRWueBest\,mm), these are reductions of \numUEtwoRed\%,
\numRWonlyRed\% and \numRWueRed\%, respectively. The mean PA-MPJPE and the MPJPE of each individual seed are also
lower than the corresponding listed values. The RW comparison has an unresolved split-version limitation:
Akada \etal~\cite{akada2024stereo} describe 86 test sequences and 7{,}936 stereo frames, whereas the current official
split used here has \numRWTestFrames frames; we have not verified the version underlying the listed RW results.

The 95\% sequence-bootstrap confidence intervals for our mean MPJPE are [\numUEtwoCIlo, \numUEtwoCIhi],
[\numRWonlyCIlo, \numRWonlyCIhi] and [\numRWueCIlo, \numRWueCIhi]\,mm.

The methods also differ in backbone, pre-training, resolution, temporal context and additional inputs. We therefore
use this table to compare complete systems. The following experiments examine performance with a common encoder
and then measure the contribution of the geometric branch by ablation.

\subsection{Comparisons with a common encoder}
\label{sec:exp_sb}

\begin{table}[t]
  \centering
  \footnotesize
  \setlength{\tabcolsep}{3.5pt}
  \begin{tabular}{@{}llccc@{}}
\toprule
 & & Val & \multicolumn{2}{c}{Test} \\
\cmidrule(lr){3-3}\cmidrule(lr){4-5}
Method & Frames & MPJPE & MPJPE & PA \\
\midrule
\multicolumn{5}{@{}l}{\emph{UnrealEgo-RW, trained on RW}} \\
EPF-device~\cite{yang2024egoposeformer} & 1 & 60.17\,{\scriptsize$\pm$1.48} & 77.85\,{\scriptsize$\pm$3.20} & 62.02\,{\scriptsize$\pm$1.22} \\
TSR-like & 8, causal & 77.97\,{\scriptsize$\pm$3.23} & 98.92\,{\scriptsize$\pm$7.27} & 77.32\,{\scriptsize$\pm$4.81} \\
\paper{}-R18 & stream & \textbf{54.73}\,{\scriptsize$\pm$0.68} & \textbf{68.76}\,{\scriptsize$\pm$1.36} & \textbf{52.78}\,{\scriptsize$\pm$1.22} \\
\midrule
\multicolumn{5}{@{}l}{\emph{UnrealEgo2, trained on UnrealEgo2}} \\
TSR-like & 8, causal & 29.76\,{\scriptsize$\pm$0.63} & 30.33\,{\scriptsize$\pm$0.55} & 27.13\,{\scriptsize$\pm$0.56} \\
\paper{}-R18 & stream & \textbf{24.96}\,{\scriptsize$\pm$0.27} & \textbf{25.57}\,{\scriptsize$\pm$0.24} & \textbf{23.85}\,{\scriptsize$\pm$0.29} \\
\bottomrule
\end{tabular}

  \caption{\textbf{Same-backbone comparison} on RW (trained on RW only) and on UnrealEgo2 (trained on UnrealEgo2 only);
    ResNet-18 + FPN at $256^2$, device-relative protocol, mm, mean\,$\pm$\,s.d.\ over three seeds. Val: validation MPJPE
    of the selected checkpoints. Test: one evaluation per frozen checkpoint. EPF-device: the official EgoPoseFormer code
    with the device origin instead of the ground-truth pelvis as projection origin; it has no UnrealEgo2 entry. TSR-like:
    our implementation following the description of TSR-Ego; not a reproduction. On UnrealEgo2, TSR-like is trained for
    12 epochs as in its paper and \paper-R18 for 3 as in our UnrealEgo2 recipe. Frames as in \cref{tab:main}.}
  \label{tab:same_backbone}
\end{table}

To assess whether the gains depend on the stronger encoder and higher resolution, we compare three systems with a
ResNet-18 + FPN encoder at $256 \times 256$ on RW (\cref{tab:same_backbone}).
\emph{EPF-device} is the official EgoPoseFormer code~\cite{yang2024egoposeformer} with its training recipe and heatmap
stage, adapted to the device-relative protocol: the public implementation uses the ground-truth pelvis of every frame as
projection origin, which we replace by the device origin given by the per-session rig.
\emph{TSR-like} is our implementation based on the description of TSR-Ego~\cite{azam2026tsrego}, with causal temporal
feature mixing and two pose-refinement layers. TSR-Ego's code is not public and several details are unspecified; our choices
are given in \cref{app:conditions}. The name denotes this local implementation, not a reproduction of TSR-Ego.
\emph{\paper-R18} is our model with the ResNet-18 encoder. It shares the heatmap-pretrained encoder, the data pipeline,
the number of epochs and the learning-rate decay with TSR-like.

On RW test, \paper-R18 reaches an MPJPE of \numSBours\,mm, \numSBdiff\,mm lower than EPF-device (95\% confidence
interval of the difference [\numSBdiffLo, \numSBdiffHi]\,mm, paired bootstrap over sequences), and a \numSBpaDiff\,mm
lower PA-MPJPE. TSR-like reaches a test MPJPE of \numSBtsr\,mm, \numSBtsrDiff\,mm higher than \paper-R18 (95\%
confidence interval [\numSBtsrDiffLo, \numSBtsrDiffHi]\,mm). The validation gap is also consistent:
TSR-like has \numTSRvalDiff\,$\pm$\,\numTSRvalDiffsd\,mm higher MPJPE than \paper-R18, across all
\numTSRvalWorse seed pairs. These results describe the specified implementation without further tuning;
\cref{sec:protocol,app:root_reference} discuss its relation to the root-relative literature.

On UnrealEgo2, TSR-like and \paper-R18 start from the same encoder, heatmap-pretrained on UnrealEgo2, but follow their
own schedules (12 epochs for TSR-like as in its paper, 3 epochs in our UnrealEgo2 recipe). \paper-R18 reaches a test
MPJPE of \numSBueOurs\,mm against \numSBueTsr\,mm, a difference of \numSBueDiff\,mm (95\% confidence interval
[\numSBueDiffLo, \numSBueDiffHi]\,mm), and a PA-MPJPE of \numSBueOursPA against \numSBueTsrPA\,mm.

The lower error on both datasets shows that the advantage over these local baselines persists with a common encoder
family and resolution. This remains a system comparison: temporal context, decoder design and parts of the training
recipe differ. The ablation below provides a more direct test of the geometric branch.

\subsection{Ablations}
\label{sec:ablation}

\begin{table}[t]
  \centering
  \footnotesize
  \setlength{\tabcolsep}{4pt}
  \begin{tabular}{@{}lcccc@{}}
\toprule
Variant & MPJPE / $\Delta$ & worse & PA / $\Delta$ & MPJVE / $\Delta$ \\
\midrule
Full model, $384^2$ input & 43.79\,{\scriptsize$\pm$0.64} & -- & 31.74 & 12.02 \\
\quad input $384^2 \to 256^2$ & +2.83\,{\scriptsize$\pm$0.78} & 3/3 & +4.15 & +0.94 \\
\quad w/o geometric branch & +0.92\,{\scriptsize$\pm$0.51} & 3/3 & +0.65 & +0.19 \\
\quad w/o trajectory sampling & $-$0.04\,{\scriptsize$\pm$0.94} & 2/3 & $-$0.25 & $-$0.07 \\
\quad w/o track prior & $-$0.54\,{\scriptsize$\pm$1.59} & 2/3 & $-$0.34 & +0.22 \\
\quad plain encoder & +0.05\,{\scriptsize$\pm$1.84} & 1/3 & $-$0.47 & $-$0.24 \\
\midrule
Full model, $256^2$ input & 46.62\,{\scriptsize$\pm$0.47} & -- & 35.89 & 12.95 \\
\quad w/o geometric branch & +2.34\,{\scriptsize$\pm$0.91} & 3/3 & +1.11 & +0.28 \\
\bottomrule
\end{tabular}

  \caption{\textbf{Ablations} on RW validation (trained on RW only, three paired seeds, mm). The rows of a full model
    give mean\,$\pm$\,s.d.\ MPJPE and mean PA-MPJPE and MPJVE; the indented rows give the change caused by the
    modification. Plain encoder: DINOv3 without the fisheye-specific components of \cref{sec:encoder}. Worse: seed pairs in
    which MPJPE increases.}
  \label{tab:ablation}
\end{table}

The geometric branch gives a consistent, modest improvement at both resolutions (\cref{tab:ablation}). Removing the
residual head, ray intersection and gate, together with their losses $\mathcal{L}_{2\mathrm{D}}$ and
$\mathcal{L}_{\mathrm{geo}}$, leaves only the learned pose offsets and increases MPJPE by
\numAblGeo\,$\pm$\,\numAblGeosd\,mm at $384^2$ and \numAblGeoLow\,$\pm$\,\numAblGeoLowsd\,mm at $256^2$.
The increase occurs in all three paired runs, and PA-MPJPE also increases. This supports the branch as a whole:
ray-consistent sampling remains in place, and the individual effects of the gate, residual head and losses are not
separated. Lowering the full model's input resolution from $384^2$ to $256^2$ increases MPJPE by
\numAblRes\,$\pm$\,\numAblRessd\,mm.

The other tested components show no consistent MPJPE improvement across seeds. Removing trajectory-aligned sampling
changes MPJPE by \numAblTrajS\,mm (higher error in \numAblTrajN of 3 pairs), and removing the track prior changes it
by \numAblPriorS\,mm (\numAblPriorN of 3 pairs); neither removal changes MPJVE consistently. Both components remain
in the evaluated configuration, which was frozen before these ablations. Causal temporal attention was not ablated.
Replacing the fisheye-specific encoder components and their heatmap stage with a plain DINOv3 encoder changes MPJPE
by \numAblEncS\,$\pm$\,\numAblEncsd\,mm (higher error in \numAblEncN of 3 pairs). The plain encoder improves
PA-MPJPE and MPJVE in \numAblEncPANeg and \numAblEncVENeg of 3 pairs, with mean changes of
\numAblEncPA and \numAblEncVE\,mm, respectively. These results do not support attributing the MPJPE gains to the
fisheye-specific encoder modifications. Our pre-specified consistency criterion and its statistical scope are given
in \cref{app:protocol}.

\subsection{Error and gate analysis}
\label{sec:analysis}

\begin{figure*}[t]
  \centering
  \genfig{\linewidth}{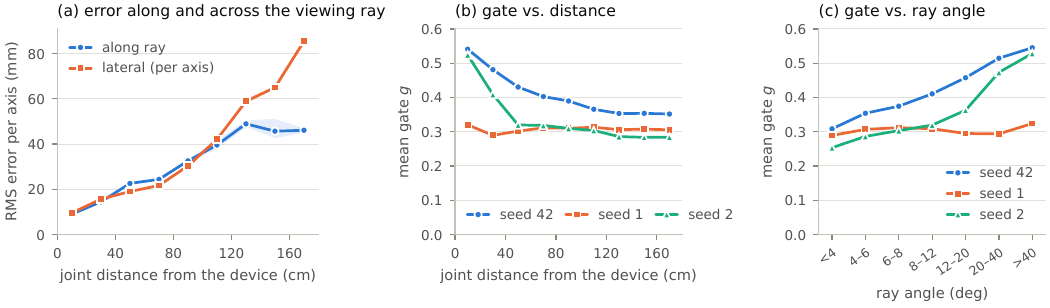}
  \caption{\textbf{Error and gate on RW validation} (final RW-only model, three seeds). (a) RMS error per axis of the final
    prediction along and across the unit vector from the device centre to the ground-truth joint, by joint distance
    (seed mean; band: range over seeds); isotropic errors would give two equal curves. (b, c) Mean gate of the last
    refinement layer by joint distance and by the angle between the two corrected viewing rays, for each seed; joints
    whose predicted anchor projects into fewer than two images are excluded.}
  \label{fig:analysis}
\end{figure*}

For the final RW-only model, validation errors increase with joint distance and become predominantly lateral for
distant joints (\cref{fig:analysis}). We decompose each joint error along and across the direction from the device
centre to the ground-truth joint, reporting RMS per axis so that isotropic errors would give equal values.
MPJPE grows from \numAnaNearMPJPE\,mm for joints
within 20\,cm to \numAnaFarMPJPE\,mm for joints 1.6--1.8\,m away. Between 40\,cm and 1\,m the along-ray component is
slightly larger than the lateral one. Beyond 1\,m the lateral component is larger and keeps growing, to
\numAnaFarLateral\,mm against \numAnaFarAlong\,mm along the ray for the farthest joints, while the along-ray component
stays between \numAnaAlongFarLo and \numAnaAlongFarHi\,mm beyond 1.2\,m. This pattern holds in all three seeds.
One untested explanation is that localising small, distorted distal joints limits angular accuracy, while body
proportions constrain their distance from the head. The failure example in \cref{app:qualitative} illustrates
large errors when the lower body is occluded in both views.

At least \numAnaValidTwo\% of predicted joint anchors project into both images. This measures image coverage;
triangulation validity additionally requires positive depth along both viewing rays (\cref{sec:geometry}). Among
these anchors, the last-layer intersection $X^{*}_j$ has a slightly lower mean error than the learned candidate in
every distance bin and seed, by up to \numAnaXsGainMax\,mm. The reported bin-mean gate values do not exceed
\numAnaGateMax; the overall means are \numAnaGateMeanA, \numAnaGateMeanB and \numAnaGateMeanC across the three seeds.
The learned update therefore retains substantial weight on average. The mean gate is lower for joints 1.6--1.8\,m away than for joints within
20\,cm by \numAnaGateDropA, \numAnaGateDropB and \numAnaGateDropC in the three seeds, and higher for ray angles above
40\textdegree{} than below 4\textdegree{} by \numAnaGateRiseA, \numAnaGateRiseB and \numAnaGateRiseC. How the gate depends
on the viewing geometry therefore differs across training runs. These descriptive averages characterise the blend
but do not establish which candidate is better for an individual joint or whether the gate is calibrated as a
reliability estimate.

\subsection{Effect of the evaluation protocol}
\label{sec:protocol}

\begin{table}[t]
  \centering
  \footnotesize
  \setlength{\tabcolsep}{3.5pt}
  \begin{tabular}{@{}lccc@{}}
\toprule
Protocol & MPJPE (own) & RA-MPJPE & PA-MPJPE \\
\midrule
Device & 77.97\,{\scriptsize$\pm$3.23} & 79.45\,{\scriptsize$\pm$4.42} & 61.45\,{\scriptsize$\pm$3.71} \\
Oracle root & 57.35\,{\scriptsize$\pm$0.74} & 57.41\,{\scriptsize$\pm$0.79} & 48.57\,{\scriptsize$\pm$0.57} \\
\midrule
Difference & $-$20.62\,{\scriptsize$\pm$3.12} & $-$22.04\,{\scriptsize$\pm$4.23} & $-$12.88\,{\scriptsize$\pm$3.71} \\
Pairs lower & 3/3 & 3/3 & 3/3 \\
\bottomrule
\end{tabular}

  \caption{\textbf{Protocol gap for one architecture} (TSR-like, ResNet-18, RW validation, three paired seeds, mm).
    Device: device-relative target and device origin, no ground-truth input. Oracle root: target relative to the
    ground-truth mid-hip, which is also the projection origin in training and at inference. Own: device-relative or
    root-relative MPJPE, respectively; RA: root-aligned. Difference: oracle root minus device, paired by seed; pairs
    lower: seed pairs in which the oracle-root run has the lower error.}
  \label{tab:protocol_gap}
\end{table}

Root-relative evaluation can change both the prediction task and the information available to the model.
TSR-Ego~\cite{azam2026tsrego} expresses poses relative to a root joint and uses that root in each camera as the
projection origin; the paper does not specify whether the root is predicted or given at inference. Root-aligned
scoring of our device-relative predictions does not provide this input. We therefore keep root-relative literature
values separate from the device-relative comparisons (\cref{app:root_reference}).

To measure the combined effect of these choices, we train the same TSR-like architecture with paired seeds under two
protocols on RW. The device protocol uses device-relative targets and the device centre as projection origin. The
oracle-root protocol uses targets relative to the ground-truth mid-hip and supplies that mid-hip as the projection
origin during both training and inference. On validation (\cref{tab:protocol_gap}), the oracle-root runs reduce MPJPE
in their own coordinate frame by \numGapOwn\,$\pm$\,\numGapOwnsd\,mm, root-aligned MPJPE by
\numGapRA\,$\pm$\,\numGapRAsd\,mm and PA-MPJPE by \numGapPA\,$\pm$\,\numGapPAsd\,mm, with the same direction in all
three seed pairs. The difference persists after root and Procrustes alignment; it is not explained by global body
translation alone. The test results follow the same direction (\cref{app:protocol_gap_test}).

This experiment measures the complete protocol change, including the training target, projection origin and scoring
frame. It neither isolates the effect of the root input at inference nor establishes how TSR-Ego was evaluated.
The size of the observed gap motivates keeping the two protocols separate when interpreting reported accuracy.

\subsection{Computational cost}
\label{sec:efficiency}

\begin{table}[t]
  \centering
  \footnotesize
  \setlength{\tabcolsep}{3pt}
  \begin{tabular}{@{}lccccc@{}}
\toprule
 & & Params & \multicolumn{2}{c}{GFLOPs} & Latency \\
\cmidrule(lr){4-5}
Model & Input & (M) & total & encoder & (ms) \\
\midrule
\paper & 384 & 35.1 & 106.9 & 97.4 & 42.8--43.7 \\
\paper{}-R18 & 256 & 13.7 & 17.9 & 13.6 & 35.4 \\
TSR-like & 256 & 12.8 & 17.4 & 13.6 & 11.2 \\
EPF-device$^{\ddagger}$ & 256 & 14.2 & 14.7 & 13.6 & 7.5 \\
\bottomrule
\end{tabular}

  \caption{\textbf{Size, computation and latency} of one streaming step (one stereo frame after an 8-frame warm-up;
    EPF-device: one single-frame forward pass). Input: image size; \paper uses DINOv3 ViT-S+/16, the other models
    ResNet-18 + FPN. GFLOPs from \texttt{torch.utils.flop\_counter}, without bilinear
    sampling and fisheye projections. Latency: median of 200 runs, batch 1, FP32, PyTorch eager, one NVIDIA H200, model
    forward only; the range covers our three final models. $^{\ddagger}$Deformable attention runs in the pure-PyTorch
    implementation of mmcv, since the precompiled CUDA kernel does not support this GPU.}
  \label{tab:efficiency}
\end{table}

\paper has \numEffParams\,M parameters and uses \numEffGflops counted GFLOPs per streaming step, with
\numEffEncShare\% in the encoder (\cref{tab:efficiency}). A step takes \numEffMsLo--\numEffMsHi\,ms on one H200 in
FP32 at batch 1. The accuracy gain comes with a latency cost: \paper-R18 and TSR-like have similar counted
computation (\numEffResNetGflops and \numEffTsrGflops GFLOPs), but take \numEffResNetMs and \numEffTsrMs\,ms,
respectively. These measurements use random weights, cover only the model forward pass, and exclude several
operations from the FLOP count; the full procedure is given in \cref{app:efficiency}. They do not establish real-time
end-to-end operation.

\section{Discussion and Conclusion}
\label{sec:conclusion}

\paragraph{Scope and limitations.}
Real-world evaluation is limited to the UnrealEgo-RW rig, and the measured computational cost leaves deployment
on a headset unresolved. External comparisons are constrained by differences in complete systems and by the unverified
RW leaderboard split correspondence. Our component analyses support the combined geometric branch but do not isolate
its individual operations or establish that the gate estimates reliability. The public test sets also informed early
development decisions. The final evaluation used fixed checkpoints, but this does not restore an unseen test set
(\cref{app:protocol}).

\paragraph{Conclusion.}
\paper connects stereo image sampling and pose refinement through shared 3D hypotheses in a ray-local frame.
Corrected-ray intersection supplies a geometric candidate, which is combined with a learned update in a causal
decoder. Local evaluations on UnrealEgo2 and UnrealEgo-RW establish the performance of the full system, while paired
ablations support the geometric branch as a whole. The remaining errors include substantial lateral error at distant
joints, motivating further investigation of image localisation alongside geometric refinement.
Our protocol analysis further shows why inference inputs and evaluation coordinates must be specified when comparing
egocentric pose estimators.

\appendix
\section{Evaluation protocol and disclosure}
\label{app:protocol}

\paragraph{Frozen evaluation.}
Before the final evaluations reported here, every test configuration was registered in a frozen list: for each entry
the run, the selected checkpoint and its SHA1 hash, the training configuration, the training record, the evaluating code
snapshot, the expected evaluation identity, the calibration files, the scorer files and the report items. The list was
recorded in our experiment log before its first use. Final result checks covered checkpoint identity, prediction
coverage for all test frames, finite predictions, and agreement between the saved predictions and scoring inputs.
The common scorer and the pipeline metric agreed to within 0.001\,mm. Each frozen checkpoint was evaluated on the test set once. Every
sequence is predicted in one causal stream from its first frame, in FP32, with the exponential-moving-average weights
and without any inference-time override. The evaluation pipeline also audits the inference inputs and records whether
any of them carries ground-truth information.

\paragraph{Statistics.}
For a configuration with seeds $s$ and test sequences $q$ we compute the three-seed mean as
$\sum_q \operatorname{mean}_s E_s(q) / \sum_q n(q)$, where $E_s(q)$ is the summed per-frame error of sequence $q$ and
$n(q)$ its number of frames; since all seeds cover the same frames, this equals the mean of the per-seed MPJPEs.
Confidence intervals resample the test sequences with replacement 10{,}000 times, frame-weighted, with one set of
resamples per dataset that is shared by all configurations, and take the 2.5 and 97.5 percentiles. The same-backbone
difference is computed on the same resamples for both models. The fitted models are held fixed, so these intervals
describe uncertainty from sampling sequences, not the full variability of training new models. Training variation is
reported separately as the sample standard deviation across three seeds. Seeds of different code bases are not paired. We state
that our results are lower than those of the leaderboard methods in a column only if our mean MPJPE is below the lowest
listed MPJPE and our mean PA-MPJPE below the lowest listed PA-MPJPE, a rule fixed before the test evaluation; we also
check every seed.

For validation ablations, our pre-specified criterion counts a component as consistently helpful for MPJPE only when
removing it increases MPJPE in all three paired runs. This is a descriptive consistency criterion, not a statistical
significance test; failure to meet it does not establish equivalence or the absence of an effect.

\paragraph{Disclosure of earlier test-set use.}
The UnrealEgo2 and UnrealEgo-RW test splits, including annotations, have been public since December 2025. During
development in September 2026, before the protocol above was fixed, we evaluated preliminary models on the test
splits: earlier versions of our model, an internal TSR-style decoder, an internal re-implementation of EgoPoseFormer,
and the official EgoPoseFormer code under the root-relative protocol with the ground-truth root as projection origin.
Several early design decisions were informed by these evaluations, among them the removal of the uncertainty module
(\cref{app:history}), and our early estimates of the difference between validation and test errors came from them. No
number from these evaluations appears in this paper. Once the protocol was fixed, the test splits were used only for
the single evaluation of each frozen checkpoint. The final results nevertheless remain subject to this prior exposure
to the public test sets.

\section{Comparison conditions}
\label{app:conditions}

\Cref{tab:conditions} lists the conditions under which the compared numbers were obtained. Items that the original papers
do not state are marked as such; we do not infer them. Our TSR-like control follows the description of
TSR-Ego~\cite{azam2026tsrego} as closely as it allows: ResNet-18 + FPN at $256^2$ with heatmap pre-training, causal
depthwise temporal mixing of the stereo feature maps, a single decoder stage with a learned initial pose, two refinement
layers in the order temporal self-attention, joint self-attention, per-view deformable stereo attention and feed-forward
layer, age embeddings, windows of 8 frames, absolute per-layer pose heads and hard field-of-view masking. The number of
mixing blocks, their kernels and dilations, the weight of the auxiliary losses and the schedule are not described in
detail; we chose them ourselves (two blocks with kernel 3 and dilations 1 and 2, auxiliary weight 0.5, our 12-epoch
schedule).

\begin{table*}[t]
  \centering
  \scriptsize
  \setlength{\tabcolsep}{3pt}
  \renewcommand{\arraystretch}{1.15}
  \newcolumntype{L}[1]{>{\raggedright\arraybackslash}p{#1}}
  \begin{tabular}{@{}L{2.0cm}L{3.2cm}L{3.0cm}L{2.9cm}L{3.5cm}L{1.5cm}@{}}
    \toprule
    Method & Encoder, input, pre-training & 3D training data & Temporal context & Extra input at inference & Numbers \\
    \midrule
    \paper (ours) & DINOv3 ViT-S+/16, $384^2$; LVD-1689M, then heatmap stage on the target training split &
      UE2: 3 epochs on UE2. RW: 12 epochs on RW. UE2 + RW: UE2 model of the same seed, then 12 epochs on RW &
      causal stream over the whole sequence; temporal attention over 8 frames, trajectory sampling over 4 past frames,
      previous-frame prior & calibration: synthetic UE2 rig; RW intrinsics of the dataset and per-session extrinsics
      fitted to training annotations & device-relative; local, frozen, 3 seeds \\
    Akada \etal~\cite{akada2024stereo} (T=5) & 2D module ResNet-18, $256^2$, ImageNet; backbones of the depth and
      heatmap feature extractors not stated & UE2: UE2. RW, UE2 + RW: UE2 pre-training, RW fine-tuning. RW only: as
      labelled on the leaderboard & stereo video, 5 frames with skip 3, last time step is predicted & depth rendered from a
      structure-from-motion reconstruction of a 4\,s window; whether the window extends beyond the current frame is not
      stated & leaderboard \\
    UnrealEgo~\cite{akada2022unrealego} & ResNet-18, $256^2$; pre-training of the leaderboard entry not stated & UE2: UE2.
      RW: UE2, then RW & single stereo frame & none & leaderboard \\
    EgoGlass$^{\dagger}$~\cite{zhao2021egoglass} & ResNet-18, $256^2$; re-implemented without the body-part segmentation
      branch & as UnrealEgo & single stereo frame & none & leaderboard \\
    TSR-Ego~\cite{azam2026tsrego} & ResNet-18 + FPN, $256^2$; dataset-specific heatmap pre-training & official training
      splits; no UE2 pre-training reported for RW & causal windows of 8 frames; test windows with stride 8 &
      fisheye calibration; the root in each camera as projection origin (UE2 pelvis, RW mid-hip); ground-truth or
      predicted root at inference not stated & root-relative, preprint \\
    EgoPoseFormer$^{*}$ \cite{yang2024egoposeformer} & official code: ResNet-18 + FPN, $256^2$; ImageNet, then heatmap
      stage & not stated & single frame & the official data loader uses the ground-truth pelvis of each frame as projection
      origin; whether the retraining kept it is not stated & root-relative, reported by~\cite{azam2026tsrego} \\
    EPF-device & official code with our RW adapter: ResNet-18 + FPN, $256^2$; ImageNet, then the official RW heatmap stage
      & RW only, official recipe: 12 epochs, batch 32 & single frame & device origin from the per-session rig as
      projection origin; no ground truth & device-relative; local, frozen, 3 seeds \\
    TSR-like & our code: ResNet-18 + FPN, $256^2$; ImageNet, then our RW or UE2 heatmap stage (shared with \paper-R18) &
      RW: 12 epochs on RW. UE2: 12 epochs on UE2 & causal windows of 8 frames & calibration as ours & device-relative;
      local, frozen, 3 seeds \\
    \paper-R18 & ResNet-18 + FPN, $256^2$; ImageNet, then our RW or UE2 heatmap stage & RW: 12 epochs on RW. UE2:
      3 epochs on UE2 (our UE2 recipe) & as \paper & calibration as ours & device-relative; local, frozen, 3 seeds \\
    \bottomrule
  \end{tabular}
  \caption{\textbf{Comparison conditions.} UE2: UnrealEgo2; RW: UnrealEgo-RW. $^{\dagger}$Re-implemented by the
    benchmark authors. $^{*}$Retrained by TSR-Ego with the official code. Local: evaluated by us on the official test
    split; frozen: under the protocol of \cref{app:protocol}.}
  \label{tab:conditions}
\end{table*}

\section{Per-session results}
\label{app:sessions}

\begin{table}[t]
  \centering
  \footnotesize
  \setlength{\tabcolsep}{4pt}
  \begin{tabular}{@{}lrcccc@{}}
\toprule
 & & \multicolumn{3}{c}{\paper} & \\
\cmidrule(lr){3-5}
Session & Frames & RW & UE2 + RW & R18 & EPF-device \\
\midrule
2023-01-17$^{\S}$ & 9,023 & 47.99 & 43.01 & 64.91 & 75.28 \\
2023-08-23 & 7,306 & 44.72 & 43.34 & 58.27 & 59.34 \\
2023-08-25 & 7,683 & 55.97 & 56.94 & 83.25 & 98.46 \\
\midrule
Upper body &  & 31.43 & 32.21 & 42.67 & 50.66 \\
Lower body &  & 67.67 & 62.92 & 94.85 & 105.04 \\
All & 24,012 & 49.55 & 47.57 & 68.76 & 77.85 \\
\bottomrule
\end{tabular}

  \caption{\textbf{RW test MPJPE per capture session} and for the upper and lower body (mm, three-seed means). RW and
    UE2 + RW: \paper trained on RW only and on UnrealEgo2 and RW; R18: \paper-R18; EPF-device as in
    \cref{tab:same_backbone}. $^{\S}$Session without training or validation data; it uses the rig of the
    chronologically nearest session.}
  \label{tab:per_session}
\end{table}

The RW test set contains one capture session that occurs in neither the training nor the validation split
(\cref{tab:per_session}). Its rig is borrowed from the nearest session. On this session, \paper trained on RW only
reaches an MPJPE of \numSessUnseen\,mm, within the range of the two other sessions (\numSessSeenLo and
\numSessSeenHi\,mm). For all models, the lower body has about twice the error of the upper body.

\section{Error and gate statistics by distance}
\label{app:analysis}

\begin{table}[t]
  \centering
  \footnotesize
  \setlength{\tabcolsep}{3pt}
  \begin{tabular}{@{}lccccccc@{}}
\toprule
 & & \multicolumn{3}{c}{final output} & \multicolumn{2}{c}{last layer} & \\
\cmidrule(lr){3-5}\cmidrule(lr){6-7}
cm & \% & MPJPE & along & lateral & $X^{*}$ & learned & gate \\
\midrule
0--20 & 6.4 & 13.5 & 9.0 & 9.6 & 13.5 & 13.6 & 0.46 \\
20--40 & 20.9 & 21.6 & 14.4 & 15.6 & 21.3 & 21.4 & 0.39 \\
40--60 & 14.6 & 28.7 & 22.6 & 19.0 & 28.1 & 29.0 & 0.35 \\
60--80 & 18.3 & 33.2 & 24.4 & 21.7 & 32.5 & 33.7 & 0.34 \\
80--100 & 5.6 & 41.4 & 32.6 & 30.3 & 40.6 & 41.9 & 0.34 \\
100--120 & 10.6 & 58.0 & 39.6 & 42.1 & 56.9 & 58.8 & 0.33 \\
120--140 & 4.9 & 74.7 & 48.9 & 58.9 & 73.5 & 75.6 & 0.31 \\
140--160 & 14.7 & 82.4 & 45.7 & 65.0 & 81.5 & 83.1 & 0.31 \\
160--180 & 4.1 & 96.4 & 46.1 & 85.4 & 96.4 & 96.7 & 0.31 \\
\bottomrule
\end{tabular}

  \caption{\textbf{RW validation errors and gate by joint distance} (final RW-only model, means over three seeds, mm).
    Along and lateral: RMS error per axis along and across the unit vector from the device centre to the ground-truth
    joint. $X^{*}$ and learned: MPJPE of the two candidates of the last refinement layer, and Gate: its mean gate, over
    the joints whose predicted anchor projects into both images.}
  \label{tab:analysis_bins}
\end{table}

\Cref{tab:analysis_bins} lists the numbers behind \cref{fig:analysis}a and the statements of \cref{sec:analysis}. The
analysis script recomputes the MPJPE of every seed from the stored validation predictions and checks it against the
evaluation record before computing the statistics.

\section{Protocol gap on the test set}
\label{app:protocol_gap_test}

The frozen test evaluation of the TSR-like decoder covers both protocols with three seeds each. Paired by seed, the
oracle-root runs have a \numGapTestOwn\,$\pm$\,\numGapTestOwnsd\,mm lower MPJPE in their own frame, a
\numGapTestRA\,$\pm$\,\numGapTestRAsd\,mm lower root-aligned MPJPE and a \numGapTestPA\,$\pm$\,\numGapTestPAsd\,mm lower
PA-MPJPE than the device-relative runs, in \numGapTestOwnN, \numGapTestRAN and \numGapTestPAN of 3 pairs, respectively. As on
validation (\cref{sec:protocol}), the difference contains the training target, the projection origin at inference and
the scoring frame, and we do not attribute it to the root input at inference.

\section{Root-relative numerical references}
\label{app:root_reference}

\begin{table}[t]
  \centering
  \footnotesize
  \setlength{\tabcolsep}{3.5pt}
  \begin{tabular}{@{}lcccc@{}}
\toprule
 & \multicolumn{2}{c}{UnrealEgo2} & \multicolumn{2}{c}{RW, trained on RW} \\
\cmidrule(lr){2-3}\cmidrule(lr){4-5}
Method & MPJPE & PA & MPJPE & PA \\
\midrule
\multicolumn{5}{@{}l}{\emph{root-relative, as reported}} \\
EgoPoseFormer~\cite{yang2024egoposeformer}$^{*}$ & 23.12 & 21.69 & 74.31 & 61.79 \\
TSR-Ego~\cite{azam2026tsrego}, 2 layers & 22.36 & 21.23 & 65.58 & 56.03 \\
TSR-Ego~\cite{azam2026tsrego}, 4 layers & -- & -- & 62.8 & 55.3 \\
\midrule
\multicolumn{5}{@{}l}{\emph{root-aligned, three-seed mean}} \\
\paper (ours) & 20.02 & 16.86 & 55.56 & 40.59 \\
\bottomrule
\end{tabular}

  \caption{\textbf{Root-relative values reported by TSR-Ego} and our root-aligned scores (test sets, mm).
    Root-relative values are quoted from~\cite{azam2026tsrego}; the root used at inference is not stated, and on
    UnrealEgo2 the root is the pelvis. Root-aligned: prediction and ground truth are each centred at their own mid-hip,
    with no ground truth at inference. PA-MPJPE (PA) does not depend on the root. $^{*}$Retrained by TSR-Ego with the
    official code. A reference, not a ranking (\cref{sec:protocol}).}
  \label{tab:cross_protocol}
\end{table}

\Cref{tab:cross_protocol} places the root-relative results reported by TSR-Ego~\cite{azam2026tsrego} alongside
root-aligned scores of our device-relative predictions. On RW, our root-aligned MPJPE is \numRArw\,mm and PA-MPJPE
is \numRWonlyPA\,mm; the reported four-layer TSR-Ego values are \numLitTsrFourRW and \numLitTsrFourRWPA\,mm.
These values are provided as numerical references. The source of TSR-Ego's root at inference is unspecified;
on UnrealEgo2 its root is the pelvis, whereas our alignment uses the mid-hip; and TSR-Ego uses 8-frame test windows
whereas our evaluation streams sequences. PA-MPJPE removes rigid transformation and scale during scoring, but cannot
remove a potential advantage from ground-truth inputs. The table therefore does not establish a ranking.

\paragraph{TSR-like under an oracle-root protocol.}
The final evaluation included three oracle-root runs of TSR-like. Before evaluation, we specified a numerical check:
whether the reported two-layer TSR-Ego value falls inside the sequence-bootstrap 95\% interval of our three-seed mean.
Our implementation reaches \numFidMean\,mm root-relative MPJPE, with an interval of
[\numFidLo, \numFidHi]\,mm, which contains the reported \numFidRef\,mm. Its PA-MPJPE is \numFidPA\,mm,
compared with \numFidPAref\,mm reported for TSR-Ego. The same local architecture reaches \numSBtsr\,mm under the
device-relative protocol (\cref{tab:same_backbone}).
The interval reflects sequence variation for the fitted models; containing a published point estimate is a descriptive
check, not evidence of statistical equivalence or a successful reproduction. Our oracle-root runs explicitly receive
the ground-truth root, while TSR-Ego's root source remains unknown, and the evaluation schedules differ as described
above.

\section{Implementation details}
\label{app:implementation}

\paragraph{Encoder.}
We use DINOv3 ViT-S+/16\footnote{timm model \texttt{vit\_small\_plus\_patch16\_dinov3.lvd1689m}.} at
$384 \times 384$, \ie $24 \times 24$ tokens per view, with stochastic depth 0.05. Cross-view blocks follow the last two transformer
blocks. The feature pyramid produces a stride-4 map with 128 channels; the auxiliary head predicts $64 \times 64$
heatmaps with a Gaussian of $\sigma = 1.5$ heatmap pixels. The heatmap stage trains the encoder on single frames at
$256 \times 256$ for 12 epochs with batch size 256, AdamW with learning rate $10^{-3}$ and weight decay
$5 \cdot 10^{-3}$, the same layer-wise decay and encoder scale as the end-to-end stage, and 500 warm-up steps.

\paragraph{Decoder.}
The decoder has width 128, 4 attention heads, feed-forward width 512, one proposal layer and three refinement layers,
with $H = 4$ heads of $K = 16$ hypotheses each. Temporal attention covers 8 frames with rotary embeddings; trajectory
sampling covers 4 past frames. The track prior is replaced by the null token with probability 0.3 and perturbed with
3\,cm Gaussian noise during training. One group of denoising queries with $\sigma = 5$\,cm is used. Samples whose
hypotheses project outside the image are replaced by the missing-view vector; for the UnrealEgo2 and UnrealEgo2 + RW
models the valid region is the fisheye image circle, for the RW-only model, which was trained with an earlier revision
of the recipe, it is the image rectangle.

\paragraph{Training and selection.}
End-to-end training uses clips of 8 consecutive frames at $384 \times 384$, batch size 16, AdamW with $\beta = (0.9,
0.999)$ and the schedule of \cref{sec:training}, bf16 mixed precision and gradient clipping at norm 5. Photometric
augmentation is shared by the two views and all frames of a clip. A checkpoint is saved after every epoch, and the one
with the lowest validation MPJPE is selected. Every run uses a single NVIDIA H200 GPU; a 12-epoch RW run takes about
five hours. EPF-device uses the official EgoPoseFormer recipe with batch size 32.

\section{Components removed during development}
\label{app:history}

Earlier versions of the decoder predicted a per-layer, per-joint uncertainty, trained with a Student-t likelihood, and
used it as an input to the gate, to the track prior and to trajectory sampling. In paired development runs with a
ResNet-18 encoder under the root-relative protocol, this module did not reduce MPJPE, and its predicted uncertainty was
over-confident: on validation, far fewer than 68\% of the errors fell inside the predicted 68\% region. We removed it
before the experiments of this paper. The same development runs did not establish an MPJPE advantage for
ray-consistent sampling alone over per-view deformable sampling, which is one reason why we do not claim the shared
sampling as a separate contribution. These development runs were also evaluated on the RW test split
(\cref{app:protocol}), and those evaluations informed the decisions.

\section{Efficiency measurement}
\label{app:efficiency}

Latency is measured for the model forward pass only, with random weights, batch size 1, PyTorch eager mode and
\texttt{cudnn.benchmark} enabled, on an otherwise idle NVIDIA H200. The inference graph has fixed shapes and no
value-dependent control flow, so FLOPs do not depend on the weights; we did not compare the latency with trained
weights. For streaming models, the state (key--value caches, trajectory history, track prior) is first warmed with 8
frames, and one further step is timed. After 30 warm-up iterations we time 200 iterations, synchronising the GPU before
and after each, and report the median. FLOPs are counted with \texttt{torch.utils.flop\_counter} on the same step,
without bilinear sampling and fisheye projections. With bf16 autocast, all models were slightly slower than in FP32 at
batch 1, so we report FP32, the precision of our evaluation.

\clearpage
\onecolumn
\section{Qualitative results and failure cases}
\label{app:qualitative}

\begin{figure}[!ht]
  \centering
  \genfig{0.70\textwidth}{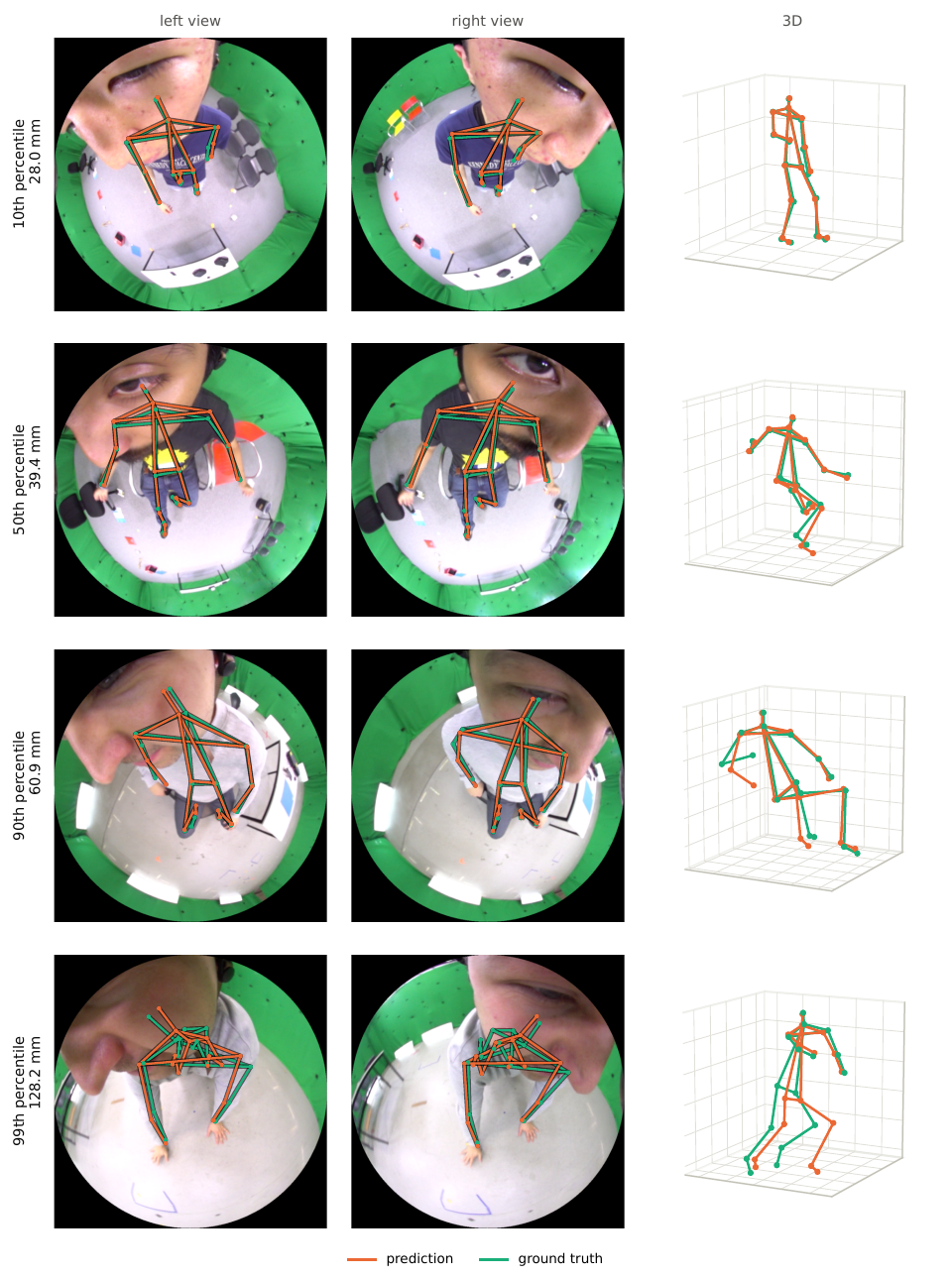}
  \caption{\textbf{Qualitative results on UnrealEgo-RW validation} (final RW-only model, seed 42). Rows: the frames whose
    per-frame error is closest to the 10th, 50th, 90th and 99th percentile, each from a different sequence; the label
    gives the frame's MPJPE. Skeletons are projected with the rig of the frame's capture session; joints that project
    outside an image are not drawn there. The lines between the neck and the hips are drawn for readability only.}
  \label{fig:qualitative}
\end{figure}

\Cref{fig:qualitative} shows validation frames chosen by a rule that we fixed before rendering any frame: for the 10th,
50th, 90th and 99th percentile of the per-frame error, the frame closest to the percentile, each from a different
sequence. The first three rows cover the typical range of errors. In the 99th-percentile frame the wearer leans forward,
the legs are hidden behind the upper body in both views, and the predicted lower body deviates from the ground truth.

{
  \small
  \begin{thebibliography}{37}
\providecommand{\natexlab}[1]{#1}
\providecommand{\url}[1]{\texttt{#1}}
\expandafter\ifx\csname urlstyle\endcsname\relax
  \providecommand{\doi}[1]{doi: #1}\else
  \providecommand{\doi}{doi: \begingroup \urlstyle{rm}\Url}\fi

\bibitem[Akada et~al.(2022)Akada, Wang, Shimada, Takahashi, Theobalt, and
  Golyanik]{akada2022unrealego}
Hiroyasu Akada, Jian Wang, Soshi Shimada, Masaki Takahashi, Christian Theobalt,
  and Vladislav Golyanik.
\newblock {UnrealEgo}: A new dataset for robust egocentric {3D} human motion
  capture.
\newblock In \emph{European Conference on Computer Vision (ECCV)}, 2022.

\bibitem[Akada et~al.(2024)Akada, Wang, Golyanik, and
  Theobalt]{akada2024stereo}
Hiroyasu Akada, Jian Wang, Vladislav Golyanik, and Christian Theobalt.
\newblock {3D} human pose perception from egocentric stereo videos.
\newblock In \emph{IEEE/CVF Conference on Computer Vision and Pattern
  Recognition (CVPR)}, 2024.

\bibitem[Akada et~al.(2025)Akada, Wang, Golyanik, and
  Theobalt]{akada2025egorear}
Hiroyasu Akada, Jian Wang, Vladislav Golyanik, and Christian Theobalt.
\newblock Bring your rear cameras for egocentric {3D} human pose estimation.
\newblock In \emph{IEEE/CVF International Conference on Computer Vision
  (ICCV)}, 2025.

\bibitem[Azam et~al.(2026)Azam, Quarles, and Desai]{azam2026tsrego}
Md~Mushfiqur Azam, John Quarles, and Kevin Desai.
\newblock {TSR-Ego}: Temporally guided stereo refinement framework for
  egocentric {3D} human pose estimation.
\newblock \emph{arXiv preprint arXiv:2607.09169}, 2026.

\bibitem[Boscolo~Camiletto et~al.(2025)Boscolo~Camiletto, Wang, Alvarado,
  Dabral, Beeler, Habermann, and Theobalt]{camiletto2025frame}
Andrea Boscolo~Camiletto, Jian Wang, Eduardo Alvarado, Rishabh Dabral, Thabo
  Beeler, Marc Habermann, and Christian Theobalt.
\newblock {FRAME}: Floor-aligned representation for avatar motion from
  egocentric video.
\newblock In \emph{IEEE/CVF Conference on Computer Vision and Pattern
  Recognition (CVPR)}, 2025.

\bibitem[Dosovitskiy et~al.(2021)Dosovitskiy, Beyer, Kolesnikov, Weissenborn,
  Zhai, Unterthiner, Dehghani, Minderer, Heigold, Gelly, Uszkoreit, and
  Houlsby]{dosovitskiy2021vit}
Alexey Dosovitskiy, Lucas Beyer, Alexander Kolesnikov, Dirk Weissenborn,
  Xiaohua Zhai, Thomas Unterthiner, Mostafa Dehghani, Matthias Minderer, Georg
  Heigold, Sylvain Gelly, Jakob Uszkoreit, and Neil Houlsby.
\newblock An image is worth 16x16 words: Transformers for image recognition at
  scale.
\newblock In \emph{International Conference on Learning Representations
  (ICLR)}, 2021.

\bibitem[Ionescu et~al.(2014)Ionescu, Papava, Olaru, and
  Sminchisescu]{ionescu2014human36m}
Catalin Ionescu, Dragos Papava, Vlad Olaru, and Cristian Sminchisescu.
\newblock {Human3.6M}: Large scale datasets and predictive methods for {3D}
  human sensing in natural environments.
\newblock \emph{IEEE Transactions on Pattern Analysis and Machine
  Intelligence}, 36\penalty0 (7):\penalty0 1325--1339, 2014.

\bibitem[Iskakov et~al.(2019)Iskakov, Burkov, Lempitsky, and
  Malkov]{iskakov2019learnable}
Karim Iskakov, Egor Burkov, Victor Lempitsky, and Yury Malkov.
\newblock Learnable triangulation of human pose.
\newblock In \emph{IEEE/CVF International Conference on Computer Vision
  (ICCV)}, 2019.

\bibitem[Jiang et~al.(2022)Jiang, Streli, Qiu, Fender, Laich, Snape, and
  Holz]{jiang2022avatarposer}
Jiaxi Jiang, Paul Streli, Huajian Qiu, Andreas Fender, Larissa Laich, Patrick
  Snape, and Christian Holz.
\newblock {AvatarPoser}: Articulated full-body pose tracking from sparse motion
  sensing.
\newblock In \emph{European Conference on Computer Vision (ECCV)}, 2022.

\bibitem[Kang and Lee(2024)]{kang2024egotap}
Taeho Kang and Youngki Lee.
\newblock Attention-propagation network for egocentric heatmap to {3D} pose
  lifting.
\newblock In \emph{IEEE/CVF Conference on Computer Vision and Pattern
  Recognition (CVPR)}, 2024.

\bibitem[Kang et~al.(2023)Kang, Lee, Zhang, and Lee]{kang2023ego3dpose}
Taeho Kang, Kyungjin Lee, Jinrui Zhang, and Youngki Lee.
\newblock {Ego3DPose}: Capturing {3D} cues from binocular egocentric views.
\newblock In \emph{SIGGRAPH Asia 2023 Conference Papers}, 2023.

\bibitem[Li et~al.(2022{\natexlab{a}})Li, Zhang, Liu, Guo, Ni, and
  Zhang]{li2022dndetr}
Feng Li, Hao Zhang, Shilong Liu, Jian Guo, Lionel~M. Ni, and Lei Zhang.
\newblock {DN-DETR}: Accelerate {DETR} training by introducing query denoising.
\newblock In \emph{IEEE/CVF Conference on Computer Vision and Pattern
  Recognition (CVPR)}, 2022{\natexlab{a}}.

\bibitem[Li et~al.(2025)Li, Yi, Liu, Gao, Ma, and Kanazawa]{li2025prope}
Ruilong Li, Brent Yi, Junchen Liu, Hang Gao, Yi Ma, and Angjoo Kanazawa.
\newblock Cameras as relative positional encoding.
\newblock In \emph{Advances in Neural Information Processing Systems
  (NeurIPS)}, 2025.

\bibitem[Li et~al.(2022{\natexlab{b}})Li, Mao, Girshick, and He]{li2022vitdet}
Yanghao Li, Hanzi Mao, Ross Girshick, and Kaiming He.
\newblock Exploring plain vision transformer backbones for object detection.
\newblock In \emph{European Conference on Computer Vision (ECCV)},
  2022{\natexlab{b}}.

\bibitem[Li et~al.(2022{\natexlab{c}})Li, Wang, Li, Xie, Sima, Lu, Yu, and
  Dai]{li2022bevformer}
Zhiqi Li, Wenhai Wang, Hongyang Li, Enze Xie, Chonghao Sima, Tong Lu, Qiao Yu,
  and Jifeng Dai.
\newblock {BEVFormer}: Learning bird's-eye-view representation from
  multi-camera images via spatiotemporal transformers.
\newblock In \emph{European Conference on Computer Vision (ECCV)},
  2022{\natexlab{c}}.

\bibitem[Li et~al.(2026)Li, Dwivedi, Maric, Chacon, Bertsch, Arcadu, Hodan,
  Ramamonjisoa, Wonka, Zhao, Kips, Keskin, Tkach, and
  Yang]{li2026egoposeformerv2}
Zhenyu Li, Sai~Kumar Dwivedi, Filip Maric, Carlos Chacon, Nadine Bertsch,
  Filippo Arcadu, Tomas Hodan, Michael Ramamonjisoa, Peter Wonka, Amy Zhao,
  Robin Kips, Cem Keskin, Anastasia Tkach, and Chenhongyi Yang.
\newblock {EgoPoseFormer v2}: Accurate egocentric human motion estimation for
  {AR/VR}.
\newblock In \emph{IEEE/CVF Conference on Computer Vision and Pattern
  Recognition (CVPR)}, 2026.

\bibitem[Liao et~al.(2024)Liao, Zhu, Wang, Hu, and
  Waslander]{liao2024mvgformer}
Ziwei Liao, Jialiang Zhu, Chunyu Wang, Han Hu, and Steven~L. Waslander.
\newblock Multiple view geometry transformers for {3D} human pose estimation.
\newblock In \emph{IEEE/CVF Conference on Computer Vision and Pattern
  Recognition (CVPR)}, 2024.

\bibitem[Lin et~al.(2022)Lin, Lin, Pei, Huang, and Su]{lin2022sparse4d}
Xuewu Lin, Tianwei Lin, Zixiang Pei, Lichao Huang, and Zhizhong Su.
\newblock {Sparse4D}: Multi-view {3D} object detection with sparse
  spatial-temporal fusion.
\newblock \emph{arXiv preprint arXiv:2211.10581}, 2022.

\bibitem[Liu et~al.(2024)Liu, Huang, Zhang, Yao, Zhang, Wan, Ye, and
  Zhou]{liu2024raydn}
Feng Liu, Tengteng Huang, Qianjing Zhang, Haotian Yao, Chi Zhang, Fang Wan,
  Qixiang Ye, and Yanzhao Zhou.
\newblock Ray denoising: Depth-aware hard negative sampling for multi-view {3D}
  object detection.
\newblock In \emph{European Conference on Computer Vision (ECCV)}, 2024.

\bibitem[Liu et~al.(2022)Liu, Wang, Zhang, and Sun]{liu2022petr}
Yingfei Liu, Tiancai Wang, Xiangyu Zhang, and Jian Sun.
\newblock {PETR}: Position embedding transformation for multi-view {3D} object
  detection.
\newblock In \emph{European Conference on Computer Vision (ECCV)}, 2022.

\bibitem[Loshchilov and Hutter(2019)]{loshchilov2019adamw}
Ilya Loshchilov and Frank Hutter.
\newblock Decoupled weight decay regularization.
\newblock In \emph{International Conference on Learning Representations
  (ICLR)}, 2019.

\bibitem[Miyato et~al.(2024)Miyato, Jaeger, Welling, and Geiger]{miyato2024gta}
Takeru Miyato, Bernhard Jaeger, Max Welling, and Andreas Geiger.
\newblock {GTA}: A geometry-aware attention mechanism for multi-view
  transformers.
\newblock In \emph{International Conference on Learning Representations
  (ICLR)}, 2024.

\bibitem[Oquab et~al.(2024)Oquab, Darcet, Moutakanni, Vo, Szafraniec, Khalidov,
  Fernandez, Haziza, Massa, El-Nouby, Assran, Ballas, Galuba, Howes, Huang, Li,
  Misra, Rabbat, Sharma, Synnaeve, Xu, Jegou, Mairal, Labatut, Joulin, and
  Bojanowski]{oquab2024dinov2}
Maxime Oquab, Timoth{\'e}e Darcet, Th{\'e}o Moutakanni, Huy Vo, Marc
  Szafraniec, Vasil Khalidov, Pierre Fernandez, Daniel Haziza, Francisco Massa,
  Alaaeldin El-Nouby, Mahmoud Assran, Nicolas Ballas, Wojciech Galuba, Russell
  Howes, Po-Yao Huang, Shang-Wen Li, Ishan Misra, Michael Rabbat, Vasu Sharma,
  Gabriel Synnaeve, Hu Xu, Herv{\'e} Jegou, Julien Mairal, Patrick Labatut,
  Armand Joulin, and Piotr Bojanowski.
\newblock {DINOv2}: Learning robust visual features without supervision.
\newblock \emph{Transactions on Machine Learning Research}, 2024.

\bibitem[Rhodin et~al.(2016)Rhodin, Richardt, Casas, Insafutdinov, Shafiei,
  Seidel, Schiele, and Theobalt]{rhodin2016egocap}
Helge Rhodin, Christian Richardt, Dan Casas, Eldar Insafutdinov, Mohammad
  Shafiei, Hans-Peter Seidel, Bernt Schiele, and Christian Theobalt.
\newblock {EgoCap}: Egocentric marker-less motion capture with two fisheye
  cameras.
\newblock \emph{ACM Transactions on Graphics (Proc. SIGGRAPH Asia)}, 2016.

\bibitem[Scaramuzza et~al.(2006)Scaramuzza, Martinelli, and
  Siegwart]{scaramuzza2006toolbox}
Davide Scaramuzza, Agostino Martinelli, and Roland Siegwart.
\newblock A toolbox for easily calibrating omnidirectional cameras.
\newblock In \emph{IEEE/RSJ International Conference on Intelligent Robots and
  Systems (IROS)}, 2006.

\bibitem[Sim{\'e}oni et~al.(2025)Sim{\'e}oni, Vo, Seitzer, Baldassarre, Oquab,
  Jose, Khalidov, Szafraniec, Yi, Ramamonjisoa, Massa, Haziza, Wehrstedt, Wang,
  Darcet, Moutakanni, Sentana, Roberts, Vedaldi, Tolan, Brandt, Couprie,
  Mairal, J{\'e}gou, Labatut, and Bojanowski]{simeoni2025dinov3}
Oriane Sim{\'e}oni, Huy~V. Vo, Maximilian Seitzer, Federico Baldassarre, Maxime
  Oquab, Cijo Jose, Vasil Khalidov, Marc Szafraniec, Seungeun Yi, Micha{\"e}l
  Ramamonjisoa, Francisco Massa, Daniel Haziza, Luca Wehrstedt, Jianyuan Wang,
  Timoth{\'e}e Darcet, Th{\'e}o Moutakanni, Leonel Sentana, Claire Roberts,
  Andrea Vedaldi, Jamie Tolan, John Brandt, Camille Couprie, Julien Mairal,
  Herv{\'e} J{\'e}gou, Patrick Labatut, and Piotr Bojanowski.
\newblock {DINOv3}.
\newblock \emph{arXiv preprint arXiv:2508.10104}, 2025.

\bibitem[Su et~al.(2024)Su, Lu, Pan, Murtadha, Wen, and Liu]{su2024roformer}
Jianlin Su, Yu Lu, Shengfeng Pan, Ahmed Murtadha, Bo Wen, and Yunfeng Liu.
\newblock {RoFormer}: Enhanced transformer with rotary position embedding.
\newblock \emph{Neurocomputing}, 568, 2024.

\bibitem[Tome et~al.(2019)Tome, Peluse, Agapito, and Badino]{tome2019xregopose}
Denis Tome, Patrick Peluse, Lourdes Agapito, and Hernan Badino.
\newblock {xR-EgoPose}: Egocentric {3D} human pose from an {HMD} camera.
\newblock In \emph{IEEE/CVF International Conference on Computer Vision
  (ICCV)}, 2019.

\bibitem[Tome et~al.(2020)Tome, Alldieck, Peluse, Pons-Moll, Agapito, Badino,
  and De~la Torre]{tome2020selfpose}
Denis Tome, Thiemo Alldieck, Patrick Peluse, Gerard Pons-Moll, Lourdes Agapito,
  Hernan Badino, and Fernando De~la Torre.
\newblock {SelfPose}: {3D} egocentric pose estimation from a headset mounted
  camera.
\newblock \emph{IEEE Transactions on Pattern Analysis and Machine
  Intelligence}, 2020.

\bibitem[Wang et~al.(2023)Wang, Luvizon, Xu, Liu, Sarkar, and
  Theobalt]{wang2023sceneego}
Jian Wang, Diogo Luvizon, Weipeng Xu, Lingjie Liu, Kripasindhu Sarkar, and
  Christian Theobalt.
\newblock Scene-aware egocentric {3D} human pose estimation.
\newblock In \emph{IEEE/CVF Conference on Computer Vision and Pattern
  Recognition (CVPR)}, 2023.

\bibitem[Wang et~al.(2024{\natexlab{a}})Wang, Cao, Luvizon, Liu, Sarkar, Tang,
  Beeler, and Theobalt]{wang2024fisheyevit}
Jian Wang, Zhe Cao, Diogo Luvizon, Lingjie Liu, Kripasindhu Sarkar, Danhang
  Tang, Thabo Beeler, and Christian Theobalt.
\newblock Egocentric whole-body motion capture with {FisheyeViT} and
  diffusion-based motion refinement.
\newblock In \emph{IEEE/CVF Conference on Computer Vision and Pattern
  Recognition (CVPR)}, 2024{\natexlab{a}}.

\bibitem[Wang et~al.(2024{\natexlab{b}})Wang, Leroy, Cabon, Chidlovskii, and
  Revaud]{wang2024dust3r}
Shuzhe Wang, Vincent Leroy, Yohann Cabon, Boris Chidlovskii, and Jerome Revaud.
\newblock {DUSt3R}: Geometric {3D} vision made easy.
\newblock In \emph{IEEE/CVF Conference on Computer Vision and Pattern
  Recognition (CVPR)}, 2024{\natexlab{b}}.

\bibitem[Wang et~al.(2021)Wang, Zhang, Cai, Yan, and Feng]{wang2021mvp}
Tao Wang, Jianfeng Zhang, Yujun Cai, Shuicheng Yan, and Jiashi Feng.
\newblock Direct multi-view multi-person {3D} pose estimation.
\newblock In \emph{Advances in Neural Information Processing Systems
  (NeurIPS)}, 2021.

\bibitem[Weinzaepfel et~al.(2022)Weinzaepfel, Leroy, Lucas, Br{\'e}gier, Cabon,
  Arora, Antsfeld, Chidlovskii, Csurka, and Revaud]{weinzaepfel2022croco}
Philippe Weinzaepfel, Vincent Leroy, Thomas Lucas, Romain Br{\'e}gier, Yohann
  Cabon, Vaibhav Arora, Leonid Antsfeld, Boris Chidlovskii, Gabriela Csurka,
  and J{\'e}r{\^o}me Revaud.
\newblock {CroCo}: Self-supervised pre-training for {3D} vision tasks by
  cross-view completion.
\newblock In \emph{Advances in Neural Information Processing Systems
  (NeurIPS)}, 2022.

\bibitem[Xu et~al.(2019)Xu, Chatterjee, Zollh{\"o}fer, Rhodin, Fua, Seidel, and
  Theobalt]{xu2019mo2cap2}
Weipeng Xu, Avishek Chatterjee, Michael Zollh{\"o}fer, Helge Rhodin, Pascal
  Fua, Hans-Peter Seidel, and Christian Theobalt.
\newblock {Mo$^2$Cap$^2$}: Real-time mobile {3D} motion capture with a
  cap-mounted fisheye camera.
\newblock \emph{IEEE Transactions on Visualization and Computer Graphics},
  2019.

\bibitem[Yang et~al.(2024)Yang, Tkach, Hampali, Zhang, Crowley, and
  Keskin]{yang2024egoposeformer}
Chenhongyi Yang, Anastasia Tkach, Shreyas Hampali, Linguang Zhang, Elliot~J.
  Crowley, and Cem Keskin.
\newblock {EgoPoseFormer}: A simple baseline for stereo egocentric {3D} human
  pose estimation.
\newblock In \emph{European Conference on Computer Vision (ECCV)}, 2024.

\bibitem[Zhao et~al.(2021)Zhao, Wei, Mahmud, and Frahm]{zhao2021egoglass}
Dongxu Zhao, Zhen Wei, Jisan Mahmud, and Jan-Michael Frahm.
\newblock {EgoGlass}: Egocentric-view human pose estimation from an eyeglass
  frame.
\newblock In \emph{International Conference on {3D} Vision (3DV)}, pages
  32--41, 2021.

\end{thebibliography}

}
\end{document}